\pdfoutput=1
\documentclass[11pt]{article}

\usepackage[utf8]{inputenc}
\usepackage{CJKutf8}
\usepackage[final]{acl}

\usepackage{times}
\usepackage{latexsym}

\usepackage[T1]{fontenc}

\usepackage[utf8]{inputenc}

\usepackage{microtype}

\usepackage{inconsolata}

\usepackage{tikz}
\usepackage{graphicx}
\usepackage{multirow}
\usepackage{amsmath}
\usepackage{amssymb}
\usepackage{wrapfig}   
\usepackage{float}
\usepackage{algorithm}
\usepackage{algorithmic}
\usepackage{mdframed}
\usepackage{tcolorbox}
\usepackage{subcaption}
\usepackage{listings}
\usepackage{lipsum}
\usepackage{array}
\usepackage{chngcntr}
\usepackage{ragged2e}
\tcbuselibrary{breakable}
\usepackage{booktabs}
\usepackage{makecell}
\usepackage{caption}
\usepackage{xcolor} 
\usepackage{pifont}  

\usepackage{booktabs}
\usepackage{multirow}
\usepackage[table]{xcolor}
\usepackage{xspace}

\usepackage[table]{xcolor} 
\definecolor{lightgray}{gray}{0.9} 
\usepackage{tcolorbox}
\tcbuselibrary{skins}
\usepackage{booktabs}
\usepackage{colortbl}
\usepackage[table]{xcolor}
\usepackage{siunitx}
\usepackage{xfp} 
\definecolor{best}{RGB}{255,235,190}   
\definecolor{second}{RGB}{220,235,255} 

\usepackage{tabularray}
\usepackage{hyperref} 
\usepackage{xstring}   
\usepackage{xurl} 
\UseTblrLibrary{booktabs}
\usepackage{longtable}
\usepackage{supertabular}

\usepackage{tabularx}
\usepackage{arydshln}

\definecolor{myGreen}{RGB}{0, 150, 0}
\definecolor{myRed}{RGB}{200, 0, 0}
\newcommand{\perf}[2]{%
    #1%
    \rlap{$
        \,_{\IfBeginWith{#2}{-}%
            {\color{myGreen}\text{\tiny{(#2)}}}%
            {\color{myRed}\text{\tiny{(#2)}}}%
        }
    $}%
}

\newtcolorbox{definitionbox}[1][]{%
  colback=blue!5,       
  colframe=blue!50!black, 
  coltitle=white,       
  colbacktitle=blue!60!black, 
  boxrule=1.5pt,                 
  rounded corners,               
  fonttitle=\bfseries,  
  enhanced,
  attach boxed title to top left={yshift=-2mm,xshift=2mm},
  boxed title style={
    rounded corners,
    borderline west={0pt}{0pt}{white}, 
    borderline east={0pt}{0pt}{white},
    borderline north={0pt}{0pt}{white},
    borderline south={0pt}{0pt}{white},
  },
  title=Definition,
  #1
}

\definecolor{thm}{RGB}{69, 53, 193}

\newcounter{assump}[section]

\newtcolorbox{assumbox}[1][]{%
  colback=blue!5,       
  colframe=blue!50!black, 
  coltitle=white,       
  colbacktitle=blue!60!black, 
  boxrule=1.5pt,                 
  rounded corners,               
  fonttitle=\bfseries,  
  enhanced,
  breakable,
  attach boxed title to top left={yshift=-2mm,xshift=2mm},
  boxed title style={
    rounded corners,
    borderline west={0pt}{0pt}{white}, 
    borderline east={0pt}{0pt}{white},
    borderline north={0pt}{0pt}{white},
    borderline south={0pt}{0pt}{white},
  },
  before upper={\refstepcounter{assump}},
  #1
}

\newtcolorbox{thmbox}[1][]{%
  colback=green!5,       
  colframe=green!50!black, 
  coltitle=white,       
  colbacktitle=green!60!black, 
  boxrule=1.5pt,                 
  rounded corners,               
  fonttitle=\bfseries,  
  enhanced,
  breakable,
  attach boxed title to top left={yshift=-2mm,xshift=2mm},
  boxed title style={
    rounded corners,
    borderline west={0pt}{0pt}{white}, 
    borderline east={0pt}{0pt}{white},
    borderline north={0pt}{0pt}{white},
    borderline south={0pt}{0pt}{white},
  },
  #1
}

\lstdefinestyle{promptstyle}{
  basicstyle=\ttfamily\small,
  columns=fullflexible,
  breaklines=true,
  breakatwhitespace=true,
  keepspaces=true,
  showstringspaces=false
}

\title{LongDS-Bench: On the Failure of Long-Horizon Agentic Data Analysis}

\author{
Kewei Xu\textsuperscript{1,3},
~Xiaoben Lu\textsuperscript{1}, 
~Shuofei Qiao\textsuperscript{1},
~Zihan Ding\textsuperscript{1}, \\
~\textbf{Haoming Xu}\textsuperscript{1}, 
~\textbf{Lei Liang}\textsuperscript{2,3},
~\textbf{Ningyu Zhang}\textsuperscript{1,3}\thanks{~Corresponding Author.}\\
\textsuperscript{1}Zhejiang University,
~\textsuperscript{2}Ant Group\\
~\textsuperscript{3}Zhejiang University - Ant Group Joint Laboratory of Knowledge Graph \\
\texttt{\{kewe1x,zhangningyu\}@zju.edu.cn} \\
}

\newcommand{\longds}{LongDS\xspace}

\begin{document}
\maketitle
\begin{abstract}

Real-world data analysis is inherently iterative, yet existing benchmarks mostly evaluate isolated or short interactive tasks, leaving agents' ability to track evolving analytical context over long horizons untested. We introduce \longds, a benchmark for long-horizon, multi-turn data analysis where agents must maintain, update, restore, and compose evolving analytical states. \longds comprises 68 tasks constructed from real-world Kaggle notebooks, spanning 2,225 turns across six domains including Geoscience, Business, and Education. Tasks are designed around state-evolution patterns (e.g., counterfactual perturbation, rollback, multi-state composition), with an average dependency span of 11.3 turns. Evaluating five state-of-the-art models, we find that the best model reaches only 48.45\% average accuracy, performance drops nearly 47 points from early to late turns, and long-horizon errors account for 52\%--69\% of failures. Further analysis shows that additional agent steps do not necessarily improve performance, suggesting that the key bottleneck is maintaining a correct analytical state rather than increasing interaction budget. We release \longds to support research on reliable long-horizon agentic data analysis\footnote{\url{https://github.com/zjunlp/DataMind}.}.

\end{abstract}

\section{Introduction}
\label{sec:introduction}

 \begin{figure}[!htbp]
    \centering
    \includegraphics[width=\linewidth]{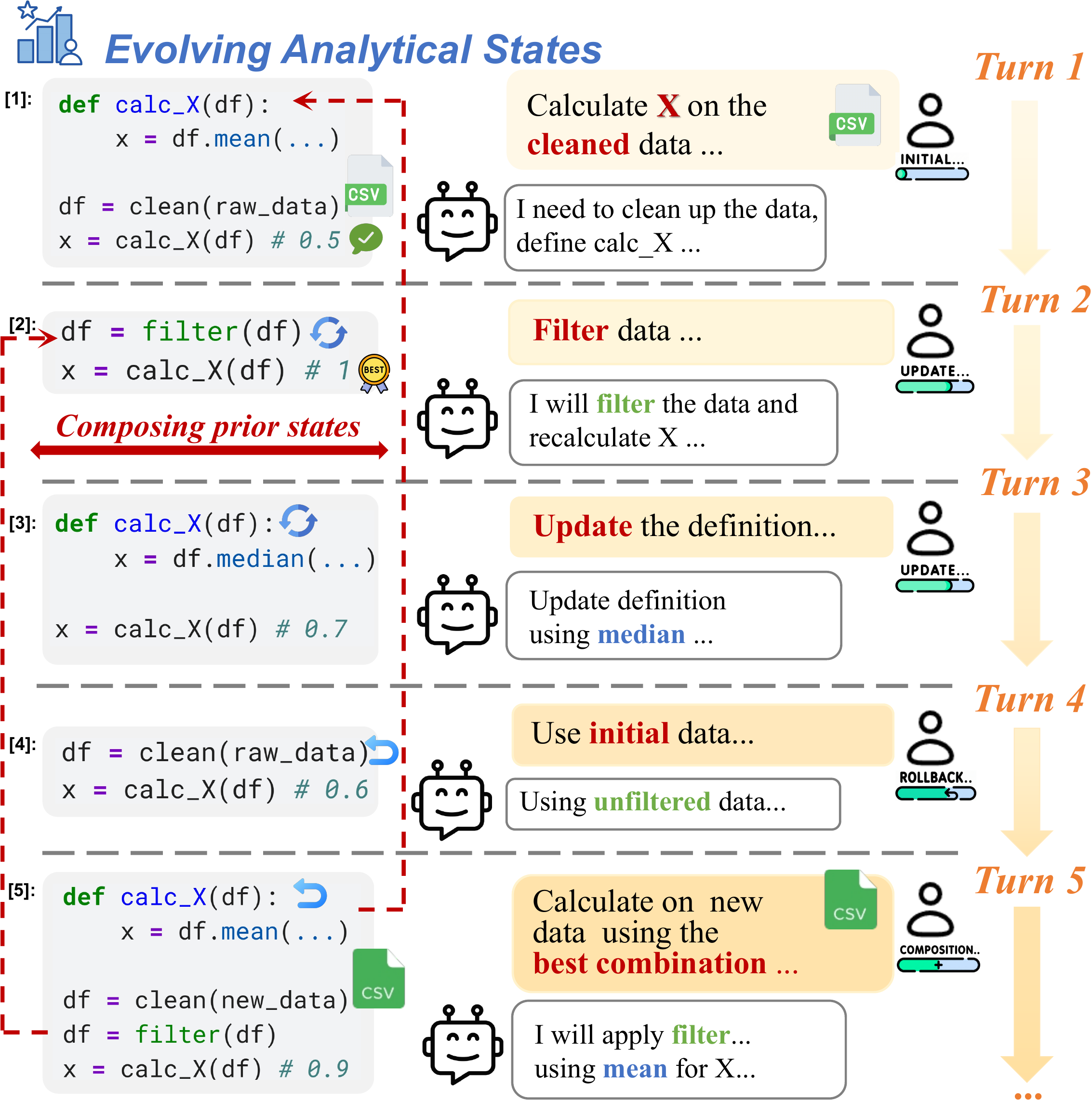} 
    \caption{
    Multi-turn, long-horizon analytical state management in \longds.
    Agents track evolving filters, definitions, and intermediate results to select the correct state for requests depending on prior turns.
    }
    \label{fig:abstract}  
    \vspace{-1.5em}
\end{figure}

Large language model (LLM) agents are increasingly used for data analysis, where they write code and execute tools to analyze data and derive insights \citep{guo2024dsagent,hong2024datainterpreter,zhang2025deepanalyze}.
However, real-world data analysis is rarely a sequence of independent, self-contained requests. 
Analytical workflows often unfold over extended persistent sessions, where scopes, metrics, assumptions, and intermediate results accumulate and shift across turns.
Handling such workflows requires maintaining an evolving analytical state for interpreting and executing each request in context.

Yet existing data analysis benchmarks provide limited evaluation of how agents manage analytical state over long horizons.
Many benchmarks focus on independent tasks in resettable environments \citep{lai2023ds1000,hu2024infiagent,jing2025dsbench,egg2025dabstep}.
Recent interactive benchmarks extend to multi-turn data analysis, but they often emphasize guided analysis completion, where the required operation is largely specified by the current turn \citep{dutta2025condabench,luo2026agentds,li2025idabench}.
As a result, they leave open whether agents can manage evolving analytical states across long dependency chains, including updating states, applying local perturbations, rolling back to earlier states, and composing multiple states.
A detailed comparison is provided in Table~\ref{tab:benchmark_comparison} and Appendix~\ref{app:detailed-related}.

\begin{figure*}[t!]  
\centering
\includegraphics[width=\textwidth]{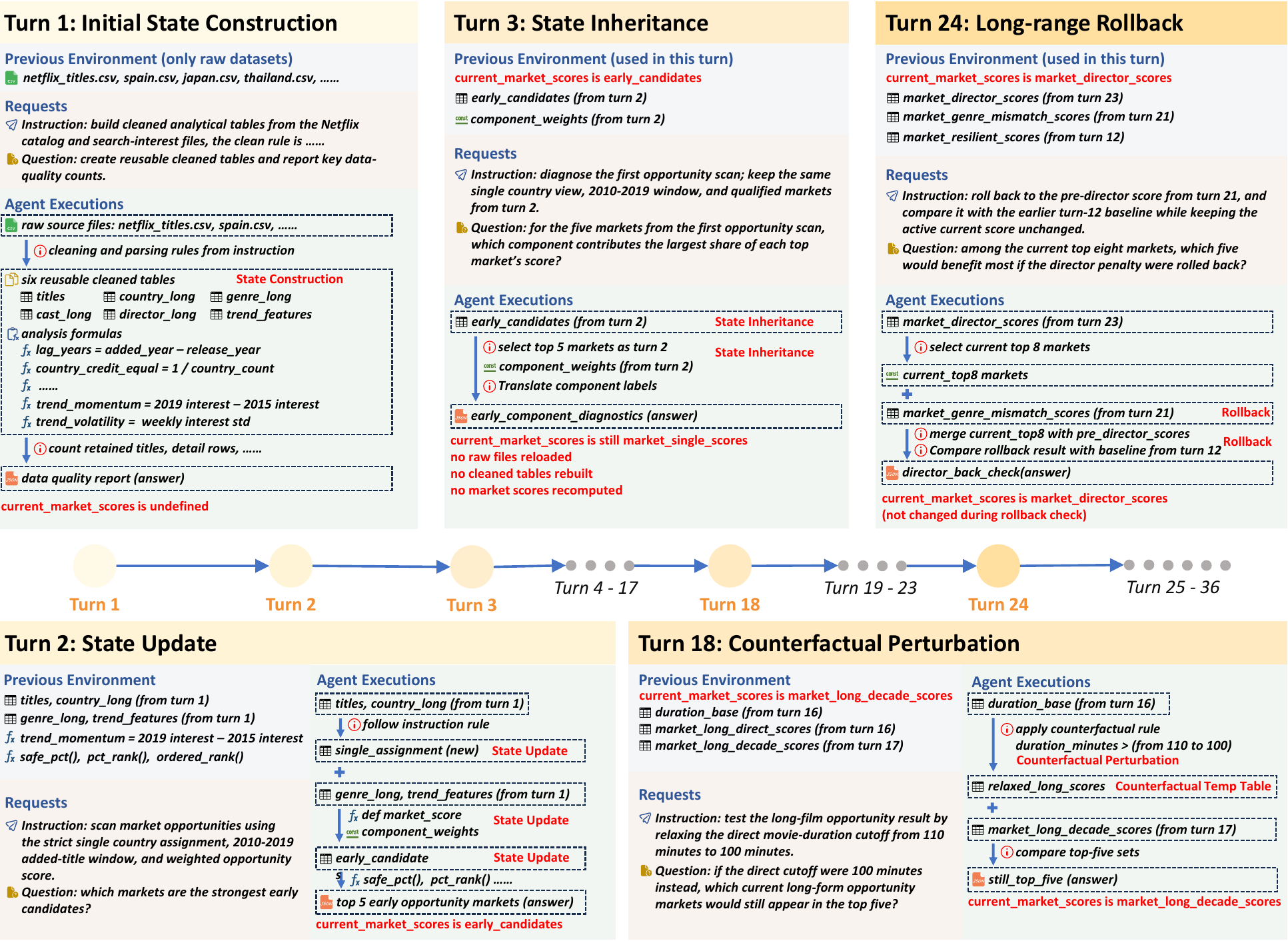}
\caption{
An example LongDS task illustrating five representative \textbf{state-evolution patterns} in a Netflix market-opportunity analysis spanning 36 turns.
\textit{Turn 1} constructs reusable analytical tables from raw files, establishing the \textbf{initial} analytical state.
\textit{Turn 2} builds on these tables to \textbf{update} the state with early market candidates.
\textit{Turn 3} \textbf{inherits} the candidates and component weights from Turn 2 to diagnose score contributors without recomputing the analysis.
\textit{Turn 18} inherits long-film scores from Turns 16--17 and applies a temporary \textbf{counterfactual perturbation} to the duration cutoff while preserving the default state.
\textit{Turn 24} uses the current top markets from Turn 23 but \textbf{rolls back} to the pre-penalty scores from Turn 12 to isolate the effect of the director-concentration penalty.
}
\label{fig:example-task}
\vspace{-0.5em}
\end{figure*}

To address this gap, we introduce \textbf{\longds}, a benchmark for evaluating long-horizon agentic data analysis over evolving analytical states.
Built from real-world Kaggle notebooks and datasets, \longds converts realistic workflows into multi-turn tasks organized around state-evolution patterns (Table~\ref{tab:state-patterns}), with long-range turn dependencies.
Figure~\ref{fig:abstract} illustrates the benchmark setting, where each task unfolds as a persistent multi-turn session and subsequent requests depend on analytical states established or updated in prior turns.
Comprising 68 tasks and 2,225 turns across six diverse application domains, including Geoscience, Business, and Education, \longds provides a challenging testbed for long-horizon analytical state management, with an average dependency span of 11.3 turns.

Our experiments reveal that long-horizon analytical state management poses a major challenge for current agents.
Across five state-of-the-art models, even the best model remains below 50\% average accuracy on \longds, with performance degrading sharply as interactions progress.
Error analysis shows that most failures are long-horizon in nature, dominated by cascading and state-management errors rather than isolated coding or reasoning mistakes.
Moreover, increasing the number of agent steps does not consistently improve accuracy, indicating that the main limitation lies in analytical state maintenance rather than interaction budget.

\definecolor{TableHeader}{HTML}{F2F2F4}
\definecolor{TableStripe}{HTML}{F7F7F9}
\definecolor{GroupText}{HTML}{00777F}
\definecolor{LongDSRow}{HTML}{EFEEFC}
\definecolor{LongDSPurple}{HTML}{6558E8}

\begin{table*}[t]
\centering
\small

\setlength{\tabcolsep}{4pt}
\renewcommand{\arraystretch}{1.08}

\begin{tabularx}{\textwidth}{
    l
    *{6}{c}
    c
    c
    >{\centering\arraybackslash}X
}






\toprule

\multicolumn{1}{l}{
    \multirow{2}{*}{\textbf{Benchmark}}
}
&
\multicolumn{1}{c}{
    \multirow{2}{*}{
        \begin{tabular}[c]{@{}c@{}}
        \textbf{Long-}\\
        \textbf{Horizon}
        \end{tabular}
    }
}
&
\multicolumn{5}{c}{
    \textbf{State Evolution Pattern}
}
&
\multirow{2}{*}{
    \textbf{\# Tasks}
}
&
\multirow{2}{*}{
    \begin{tabular}[c]{@{}c@{}}
    \textbf{Avg.}\\
    \textbf{Turns}
    \end{tabular}
}
&
\multirow{2}{*}{
    \textbf{Turn Purpose}
}
\\

\cline{3-7}

\multicolumn{1}{c}{}
&
\multicolumn{1}{c}{}
&
\textbf{Init}
&
\textbf{Inh}
&
\textbf{Upd}
&
\textbf{CF}
&
\textbf{RB}
&
&
&
\\

\midrule

\addlinespace[0pt]

\multicolumn{10}{l}{
    \textcolor{GroupText}{\footnotesize\textit{Single-Turn}}
} \\[-1pt]

\rowcolor{TableStripe}
DABStep~\citep{egg2025dabstep}
& $\times$ & $\times$ & $\times$ & $\times$ &$\times$ & $\times$
& 450
& N/A
& N/A \\

AgentDS~\citep{luo2026agentds}
& $\diamond$ & $\times$ & $\times$ & $\times$ & $\times$ & $\times$
& 17
& N/A
& N/A \\

\rowcolor{TableStripe}
LongDA~\citep{li2026longdabenchmarkingllmagents}
& $\checkmark$ & $\times$ & $\times$ & $\times$ & $\times$ & $\times$
& 505
& N/A
& N/A \\

\midrule

\multicolumn{10}{l}{
    \textcolor{GroupText}{\footnotesize\textit{Multi-Turn}}
} \\[0pt]

\rowcolor{TableStripe}
ConDABench~\citep{dutta2025condabench}
& $\times$ & $\diamond$ & $\diamond$ & $\times$ & $\times$ & $\times$
& 1,420
& $\sim$1.81
& Task clarification \\

IDA-Bench~\citep{li2025idabench}
& $\diamond$ & $\diamond$ & $\diamond$ & $\diamond$
& $\times$ & $\times$
& 25
& $\sim$9.80
& Task guidance \\

\midrule

\rowcolor{LongDSRow}
\textcolor{LongDSPurple}{\textbf{LongDS}}
& $\checkmark$
& $\checkmark$
& $\checkmark$
& $\checkmark$
& $\checkmark$
& $\checkmark$
& \textbf{68\textsuperscript{\dag}}
& \textbf{$\sim$32.7}
& \textbf{New analytical request} \\

\bottomrule
\end{tabularx}

\vspace{0pt}

\caption{
Comparison of \textbf{LongDS} with existing data analysis benchmarks.
\textbf{LongDS} targets the \textbf{long-horizon} management of
\textbf{evolving analytical states} over extended multi-turn analysis.
Init., Inh., Upd., CF, and RB denote initial construction, inheritance,
update, counterfactual, and rollback, respectively.
$\diamond$ denotes implicit or partial coverage.
$\dagger$ The 68 LongDS tasks span a total of \textbf{2,225} turns,
each corresponding to a concrete data analysis request.
}
\label{tab:benchmark_comparison}

\end{table*}

In summary, our contributions are threefold:
\begin{itemize}
    \item We formulate long-horizon agentic data analysis as analytical state management, covering initial construction, inheritance, updates, counterfactual perturbations, rollbacks, and multi-state composition.
    \item We introduce \longds, a realistic benchmark constructed from real-world workflows, comprising 68 tasks and 2,225 turns with long-range state dependencies.
    \item We provide a systematic evaluation of strong proprietary and open-source models, revealing substantial performance degradation over long trajectories and failures dominated by cascading and state-management errors.
\end{itemize}

\begin{figure*}[t!]
\centering
\resizebox{1\textwidth}{!}{
\includegraphics{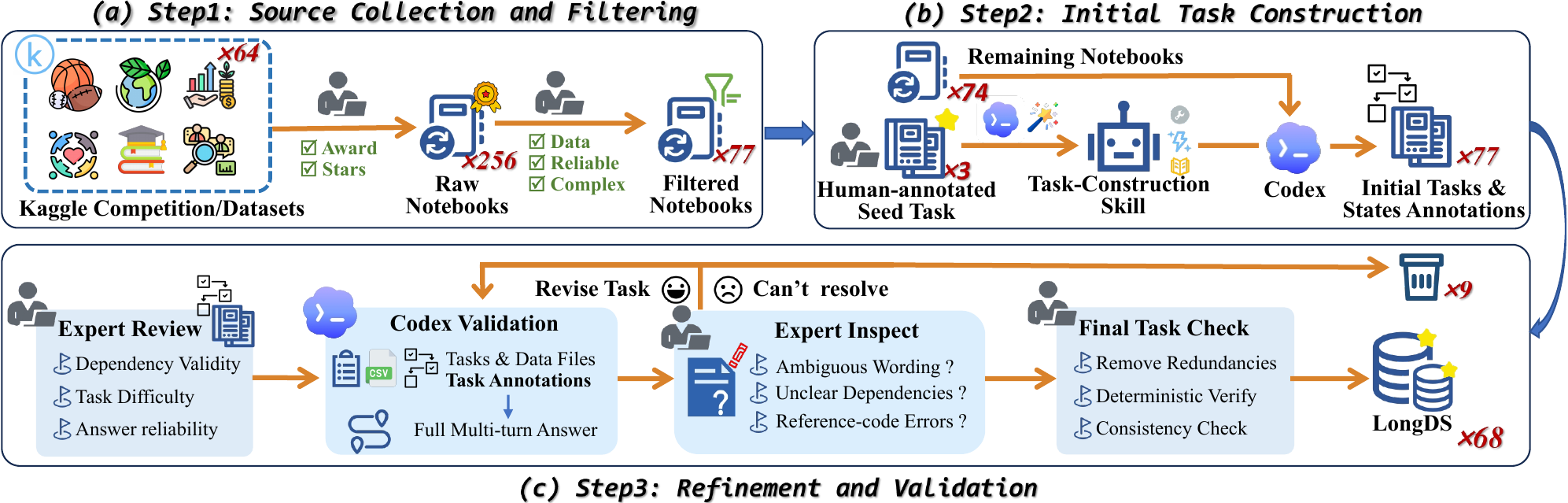}}
\caption{
Task curation pipeline of \longds: 
(a) source collection and filtering, 
(b) initial state-annotated task construction, and 
(c) refinement through expert review, Codex-based validation, and consistency checks.
}
\label{fig:benchmark}
\vspace{-1em}
\end{figure*}

\section{Preliminary}
\label{sec:preliminary}

A multi-turn data analysis task consists of a sequence of user requests over a collection of data files, carried out in a persistent executable environment such as a Jupyter notebook.
Formally, a task is defined as:
\[
\small
\mathcal{T} = (\mathcal{D}, E_0, U)
\]
where $\mathcal{D}$, $E_0$, and $U=(u_1,\ldots,u_T)$ denote the data files, the initial executable environment, and the sequence of user requests, respectively.

At turn $t$, an agent receives the current request $u_t$, interaction history $H_{<t}$, and current environment state $E_{t-1}$.
Here, $H_{<t}$ denotes the interaction history before turn $t$, including previous user requests, agent responses, and executed analysis steps.
The agent then performs analysis and returns a response $y_t$, resulting in an updated environment state $E_t$.
Unlike isolated data analysis tasks, the environment is not reset between turns, allowing intermediate code states and results to persist.

The target response at turn $t$ is determined by the current request $u_t$, the current environment state $E_{t-1}$, and the analytical context accumulated in $H_{<t}$, such as prior scopes, definitions, and assumptions.
The goal of a multi-turn data analysis agent is to produce a sequence of target responses:
\[
\small
Y = (y_1, \ldots, y_T)
\]
This setting captures long-horizon data analysis, where later requests may depend on analytical states that are inherited, revised, temporarily perturbed, or restored from many turns earlier.

\section{LongDS Benchmark}
\label{sec:benchmark}

\subsection{Design Principles}
\label{sec:design-principles}

\longds evaluates data-analysis agents' ability to reason over \textbf{evolving analytical states} across long-horizon interactions, including scopes, definitions, assumptions, and intermediate results.
A central challenge is that the valid analytical state is not static. 
Across a multi-turn trajectory, user requests may introduce new analytical objects, inherit existing ones, revise previous definitions, temporarily perturb assumptions, or restore earlier versions of the analysis.  
We therefore construct tasks around the state-evolution patterns summarized in Table~\ref{tab:state-patterns}.

These patterns differ in how they affect the active analytical state: \textbf{updates} overwrite the default state, \textbf{counterfactual perturbations} apply only locally, \textbf{rollbacks} answer the current request under an earlier anchored state, and \textbf{multi-state composition} requires combining multiple states.
\textbf{Inheritance} is included for clarity, but it is treated as the default persistence mechanism rather than a separately annotated category in benchmark statistics.
Figure~\ref{fig:example-task} provides a representative example of such long-horizon state evolution in a 36-turn Netflix market-opportunity analysis task.

\subsection{Task Curation}
\label{sec:task-curation}

We construct \longds from real-world data analysis notebooks to build long-horizon interactive tasks where later requests depend on analytical states established and revised across realistic workflows.
The construction process consists of three steps as shown in Figure~\ref{fig:benchmark}.

\paragraph{Source Collection and Filtering.}
We curate raw notebooks from established Kaggle analytics competitions and highly upvoted public datasets.
Unlike prediction-focused competitions, these sources capture full analysis workflows rather than isolated prediction tasks, providing realistic chains for constructing long-horizon tasks.
This process yields an initial pool of 64 competitions and datasets.
For each source, we select four notebooks from winning submissions or highly upvoted public notebooks, resulting in 256 high-quality raw notebooks spanning diverse domains.

We then manually execute and inspect each selected notebook, filtering them based on data accessibility, execution reliability, and computational feasibility.
We fix minor execution issues or obvious code errors when they do not alter the intended analysis, and remove notebooks with insufficient analytical depth for long-horizon task construction.
After filtering and repair, we retain 36 competitions and datasets, comprising 77 executable filtered notebooks, as shown in Figure~\ref{fig:benchmark}(a).

\paragraph{Initial Task Construction.}
We first manually construct three seed tasks from representative filtered notebooks, guided by three principles: preserving the original analytical thread, formulating turns as quantitatively evaluable questions rather than visualization requests, and designing realistic long-range dependencies over evolving analytical states.
We decompose each notebook into analysis clusters, i.e., groups of cells with shared computational objectives.
Within each analysis cluster, we identify key analytical objects, such as data scopes, derived definitions, metrics, and intermediate results, and convert them into reusable analytical anchors.
We then design state-dependent turns that reuse these anchors within and across analysis clusters to create long-range dependencies, following the \textbf{state-evolution patterns} in Section~\ref{sec:design-principles}.

Based on the three manually constructed seed tasks, we use \textit{Codex} \citep{openai_codex}, a coding agent equipped with \textit{skill-creator}, to build a reusable \textit{task-construction skill}.
Specifically, we provide \textit{Codex} with detailed design principles, three original notebooks, and corresponding converted seed tasks as paired demonstrations.
The resulting skill encodes the construction procedure, including decomposition into analysis clusters, anchor identification, state-dependent turn design, and task annotations.
We then use \textit{Codex} with this \textit{task-construction skill} to convert the remaining filtered notebooks into initial tasks.
Each constructed task contains a long-horizon sequence of turns, where each turn includes a user request, executable reference code, a reference answer, state-evolution labels, and inter-turn dependency annotations.
Together with the three seed tasks, this process yields 77 initial tasks, as shown in Figure~\ref{fig:benchmark}(b).

\begin{figure}[t!]
\centering
\includegraphics[width=\linewidth]{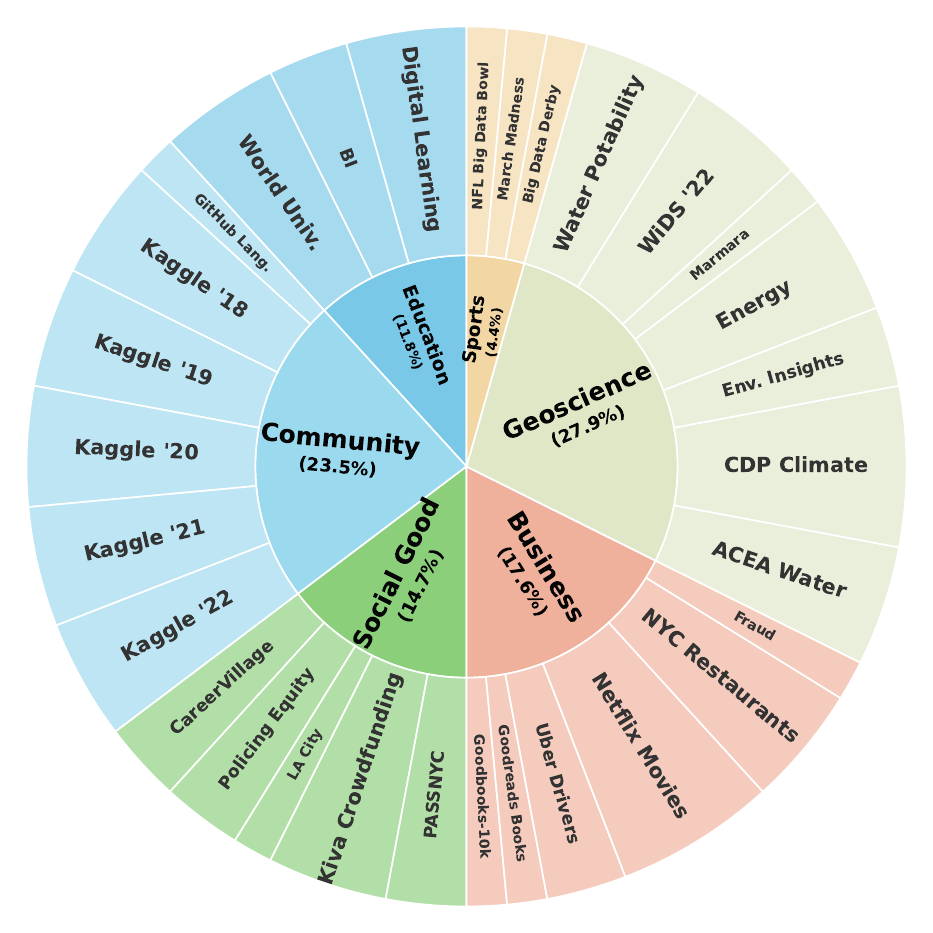}
\caption{
Domain and task distribution of \longds. 
The inner ring shows application domains, while the outer ring shows source datasets and Kaggle competitions, with sector size proportional to the number of long-horizon analysis tasks.
}
\label{fig:domain-distribution}
\vspace{-1em}
\end{figure}

\paragraph{Refinement and Validation.}
To ensure task quality, we conduct a three-stage refinement and validation process, as shown in Figure~\ref{fig:benchmark}(c).

\textbf{Stage 1: Expert Review.}
Expert reviewers with graduate-level training in data science and NLP manually inspect each task against three quality criteria: dependency validity, task difficulty, and answer reliability.
The generated task annotations are first reviewed to verify that each turn depends on the correct prior states and that these dependencies are necessary for the intended analysis.
Weakly dependent or overly simple turns are revised to require richer use of analytical states and longer-range dependencies.
The reference code for the full task is then rerun to verify that it executes successfully and produces reproducible answers.

\textbf{Stage 2: Annotation-Guided Validation.}
We then use \textit{Codex} for annotation-guided validation by providing all requests, data files, and task annotations for each reviewed task.
This stage serves as an internal consistency check, using annotations to verify consistency among task specifications, dependencies, and reference answers.
We compare \textit{Codex} outputs with the reference answers and manually inspect mismatches to identify task-quality issues, such as ambiguous wording, missing information, incorrect annotations, or reference code errors.
When such issues are found, we revise the task and rerun validation until no further task-quality issues are identified.
Tasks whose ambiguity, weak dependency, or reliability issues cannot be resolved are discarded, resulting in 68 final tasks.

\textbf{Stage 3: Final Task Check.}
Following the validation, we selectively remove redundant information from the final requests (e.g., restatements of earlier filters or metric definitions) so that long-range dependencies are not made overly explicit.
After this removal step, we verify that each answer remains uniquely derivable from the final task specification and the provided data.
Finally, we conduct a consistency check to ensure that the final requests, executable reference code, reference answers, and task annotations remain aligned.

\subsection{Benchmark Statistics}
\label{sec:benchmark-statistics} 


Following the task curation pipeline and manual quality validation, \longds contains \textbf{68} long-horizon data analysis tasks spanning six application domains: Sports, Geoscience, Business, Social Good, Education, and Community, as shown in Figure~\ref{fig:domain-distribution}.
Together, these tasks comprise \textbf{2,225} turns in total, with an average of 33 turns per task.

As shown in Appendix Figure~\ref{fig:state-statistics}, state-evolution patterns are frequent and diverse, with each task averaging 5.8 rollback turns and 8.6 multi-state composition turns, alongside frequent updates and counterfactual perturbations.
Dependency structure is similarly demanding, with an average breadth of 2.9 dependencies per turn and an average span of 11.3 turns, confirming that \longds requires agents to track and compose analytical states across long interaction histories.
Detailed task-level macro statistics are provided in Appendix~\ref{app:benchmark-statistics}.

We further examine potential training-data contamination arising from the use of public Kaggle notebooks.
\textbf{Directed n-gram containment} indicates limited verbatim overlap with the source notebooks, while a \textbf{no-data memorization probe} yields near-zero performance without access to the underlying data.
Full results are provided in Appendix~\ref{app:contamination}.

\subsection{Evaluation Protocol}
\label{sec:evaluation-protocol}
Each turn is paired with executable reference code and a structured reference answer, enabling reproducible evaluation.
We use DeepSeek-V4-Pro~\citep{deepseekai2026deepseekv4} as the judge model~\citep{liu2023geval,zheng2023judgingllm,kim2024prometheus} to assess whether an agent's answer is semantically and numerically consistent with the reference answer, without constraining the output format.
For turns involving model training, a small numerical tolerance is permitted to account for non-determinism.

Formally, the score $s_{i,j}$ for the $j$-th turn in the $i$-th task is defined as:
\begin{equation}
    \small
    s_{i,j} = 
    \begin{cases} 
        1, & \text{if consistent with the reference}, \\ 
        0, & \text{otherwise}.
    \end{cases}
\end{equation}
For \longds containing $M$ tasks where the $i$-th task consists of $N_i$ turns, the average score $S_{\text{avg}}$ is defined as the macro-average of task-level scores:
\begin{equation}
    \small
    S_{\text{avg}} = \frac{1}{M} \sum_{i=1}^{M} \left( \frac{1}{N_i} \sum_{j=1}^{N_i} s_{i,j} \right).
\end{equation}

To validate the reliability of this automated evaluation protocol, we conduct a blind human audit, finding strong agreement between human and LLM judgments, with 93.11\% agreement and Cohen's $\kappa$ of 0.8623~\citep{bavaresco-etal-2025-llms,chiang-lee-2023-large,Landis1977TheMO}. 
Further details are provided in Appendix~\ref{app:judge-validation}.

\begin{table*}[htbp!]
\centering

\resizebox{\textwidth}{!}{%
\begin{tabular}{@{}lcccccc|cc@{}}
\toprule
\textbf{Model} & \multicolumn{1}{c}{\textbf{Education}} & \multicolumn{1}{c}{\textbf{Community}} & \multicolumn{1}{c}{\textbf{Social Good}} & \multicolumn{1}{c}{\textbf{Business}} & \multicolumn{1}{c}{\textbf{Geoscience}} & \multicolumn{1}{c|}{\textbf{Sports}} & \multicolumn{1}{c}{\textbf{Avg Score}} & \multicolumn{1}{c}{\textbf{Avg Step}} \\ \midrule
\multicolumn{9}{c}{\textbf{Proprietary Models}} \\ \midrule
GPT-5.4            & \cellcolor{cyan!18}\textbf{77.92} & \cellcolor{cyan!8}65.32 & \cellcolor{cyan!8}36.80 & \cellcolor{cyan!8}28.40 & 28.90 & 10.52 & \cellcolor{cyan!8}43.50 & 68.57 \\
Claude-4.6-Sonnet  & \cellcolor{cyan!8}77.29 & 54.64 & 36.10 & 25.54 & \cellcolor{cyan!8}31.92 & 19.76 & 41.56 & 170.04 \\
Gemini-3.1-Pro      & 58.03 & \cellcolor{cyan!18}\textbf{69.54} & \cellcolor{cyan!18}\textbf{41.73} & \cellcolor{cyan!18}\textbf{33.59} & \cellcolor{cyan!18}\textbf{42.20} & \cellcolor{cyan!8}31.85 & \cellcolor{cyan!18}\textbf{48.45} & 117.82 \\ \midrule
\multicolumn{9}{c}{\textbf{Open-source Models}} \\ \midrule
Kimi-K2.6          & 64.98 & 60.62 & 31.29 & 20.99 & 28.83 & \cellcolor{cyan!18}\textbf{32.85} & 39.72 & 115.41 \\
DeepSeek-V4-Pro   & 61.36 & 49.47 & 32.41 & 17.06 & 16.60 & 15.82 & 31.97 & 133.12 \\ \bottomrule
\end{tabular}
}
\caption{
\textbf{Main results across six domains in \longds. }
Scores are macro-averaged over task-level accuracies (\%), with each task accuracy computed over its turns. 
Avg Step denotes average agent steps across all tasks. 
\protect\colorbox{cyan!18}{\textbf{Best}} and 
\protect\colorbox{cyan!8}{second-best} scores are highlighted, excluding Avg Step.
}
\label{tab:main_results}
\end{table*}

\begin{figure*}[t!]
\centering
\includegraphics[width=\textwidth]{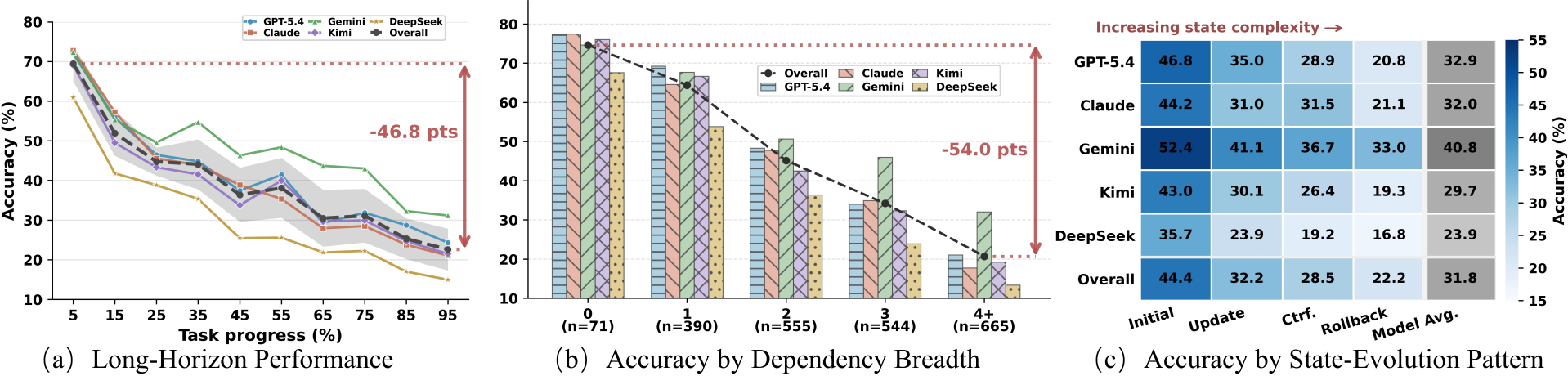}
\caption{
\textbf{Long-horizon performance degradation in \longds.}
Accuracy drops across three increasing demands: 
(a) later task progress, averaged within each 10\% progress interval; 
(b) larger dependency breadth, with $n$ denoting the number of turns per group; 
and (c) more complex state-evolution patterns.
}
\label{fig:analysis-all}
\vspace{-1em}
\end{figure*}

\section{Experiments}
\label{sec:experiments}

\subsection{Experiment Settings}

We conduct our experiments using the DSGYM framework~\citep{nie2026dsgym}.
The data analysis agent employs a ReAct-style strategy~\citep{yao2023react}, generating reasoning traces and Python code executed in a persistent Jupyter Notebook kernel. 
Final answers are extracted from the agent's response to facilitate automated semantic evaluation.

We evaluate a diverse set of state-of-the-art LLMs on \longds, including GPT-5.4~\citep{OpenAI-GPT54}, Gemini-3.1-Pro~\citep{GoogleGemini31Pro}, Claude-4.6-Sonnet~\citep{AnthropicClaudeSonnet46}, DeepSeek-V4-Pro, and Kimi-K2.6~\citep{KimiK26TechBlog}. 
To accommodate complex multi-step reasoning and iterative debugging within long-horizon tasks, we set the maximum number of reasoning-action steps to 40 per turn.
Further implementation details are provided in Appendix~\ref{app:experiment-details}.
\subsection{Main Results}

\noindent \textbf{Overall Performance.} Table~\ref{tab:main_results} presents model results across the six domains in \longds.
Overall, even the best-performing model remains below 50\% average accuracy, indicating that \longds poses a substantial challenge for current LLM agents.
\textbf{Gemini-3.1-Pro} achieves the highest average score, reaching \textbf{48.45}, and leads in Community, Social Good, Business, and Geoscience.
GPT-5.4 and Claude-4.6-Sonnet follow with average scores of 43.50 and 41.56, respectively.
Notably, \textbf{Kimi-K2.6} achieves an average score of 39.72, comparable to those of some proprietary models.
We also evaluate \textit{Codex} on a sampled subset of tasks, with details provided in Appendix~\ref{app:codex_results}.
On this subset, \textit{Codex} improves over the ReAct-based Gemini-3.1-Pro baseline by 4.38 points, suggesting that a stronger agent does not fully resolve the long-horizon state-management challenge.


\paragraph{Domain Variance.}
Performance varies substantially across domains.
Models generally achieve higher scores in Education, while lower performance is observed in Geoscience and Business, which often involve more complex feature engineering and long-horizon statistical reasoning.
Bootstrap 95\% confidence intervals further reveal greater uncertainty for smaller domains, as detailed in Appendix~\ref{app:statistical}.
Overall, these cross-domain differences suggest that no single model consistently manages long-horizon analytical state across diverse domains.

\paragraph{Degradation in Long-Horizon State Tracking.}
Model performance degrades as long-horizon state tracking becomes more demanding.
(1) \textit{Accuracy decreases as tasks progress}. 
Figure~\ref{fig:analysis-all}(a) shows a nearly 47 percentage-point drop between the first and last 10\% progress intervals after normalizing turn positions within each task.
This decline suggests that agents struggle as analytical states accumulate.
(2) \textit{Dependency structure introduces an additional bottleneck}. 
Accuracy drops sharply as dependency breadth increases in Figure~\ref{fig:analysis-all}(b), with a similar decline under longer dependency spans in Figure~\ref{fig:appendix-span}.
(3) \textit{Performance declines as turns involve more complex state transitions.}
Figure~\ref{fig:analysis-all}(c) shows a clear decline from Initial to Update, Counterfactual, and Rollback requests.
This suggests that agents handle state construction relatively well, but struggle increasingly to revise, temporarily perturb, and restore analytical states.
Together, these results show that \longds stresses agents' ability to maintain, revise, restore, and compose analytical states across extended data-analysis trajectories.

\section{Deep Analysis}
\label{sec:analysis}

\subsection{Efficiency and Performance Trade-off}

As shown in Table~\ref{tab:main_results}, GPT-5.4 uses fewer agent steps than the other models, whereas Claude-4.6-Sonnet uses the most.
Figure~\ref{fig:analysis}(a) further compares model performance with resource consumption, measured by agent steps and trajectory tokens.
Gemini-3.1-Pro achieves the highest accuracy, while GPT-5.4 obtains the best cost-normalized efficiency due to its lower step and token usage. 
Claude-4.6-Sonnet uses the most steps but does not achieve the best accuracy, suggesting that more interaction budget does not necessarily improve long-horizon performance.

Domain-level results in Figure~\ref{fig:analysis}(b) show similar trends, with GPT-5.4 exhibiting strong resource efficiency across domains.
Overall, these results suggest that \textbf{longer analysis is not inherently better} for long-horizon data analysis.
A key factor is whether the model can \textbf{maintain a correct analytical state} throughout the trajectory, as additional steps may introduce opportunities for state drift.

\subsection{Error Analysis}
\label{subsec:error_analysis}
\paragraph{Error Categorization.}
We first classify incorrect turns into general error types following DSGYM, including instruction following, statistical and domain reasoning, coding, and planning errors.
To capture failures specific to long-horizon interactions, we additionally define three long-horizon error types:
\textbf{Context Memory Error} refers to failures in recalling or using relevant historical information.
\textbf{State Management Error} occurs when models select, update, or restore the wrong analytical state.
\textbf{Cascade Error} denotes cases where the current turn is locally correct but fails due to incorrect intermediate states propagated from earlier turns.
We use \textit{Codex} to annotate the incorrect turns based on this taxonomy, and validate reliability through human auditing (Cohen's $\kappa$=0.75).
Further details are provided in Appendix~\ref{app:error_analysis}.

\begin{figure*}[!t]
\centering
\includegraphics[width=\textwidth]{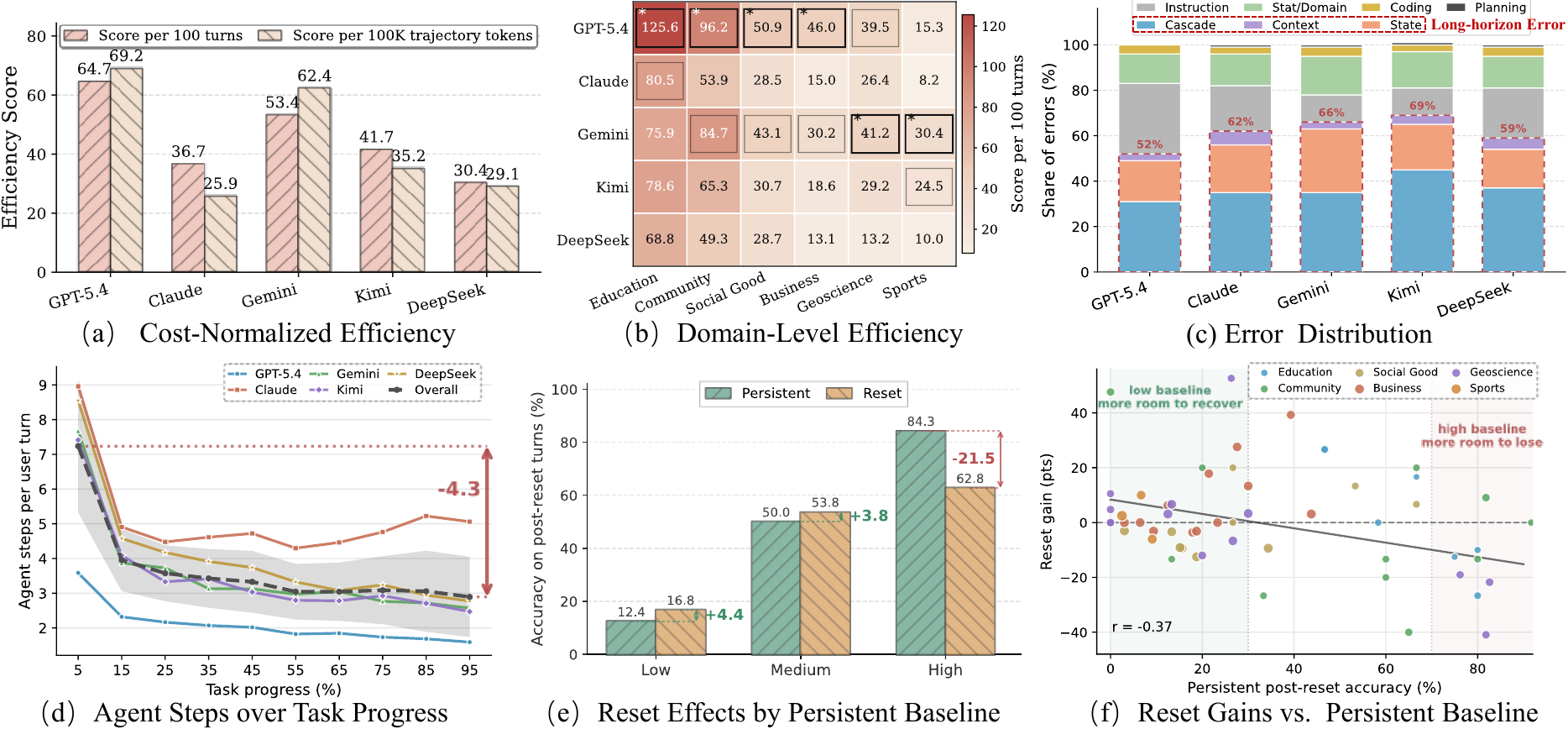}
\caption{
\textbf{Diagnosing the state-maintenance bottleneck in \longds.}
\textbf{(a) Cost-Normalized Efficiency:} Higher interaction cost does not necessarily yield better performance.
\textbf{(b) Domain-Level Efficiency:} Efficiency varies across domains and models.
\textbf{(c) Error Type Distribution:} Long-horizon errors, including Cascade, Context, and State errors, dominate failures.
\textbf{(d) Agent Steps Over Task Progress:} Agents take fewer steps as tasks progress.
\textbf{(e) Reset Effects by Persistent Baseline:} Reset helps weak persistent states but hurts strong ones.
\textbf{(f) Reset Gains vs. Persistent Baseline:} Reset exposes a recovery--continuity trade-off.
}
\label{fig:analysis}
\vspace{-1em}
\end{figure*}

\paragraph{Dominance of Long-Horizon Errors.}
Figure~\ref{fig:analysis}(c) illustrates the error distribution across models.
Long-horizon errors account for the majority of failures, ranging from 52\% for GPT-5.4 to 69\% for Kimi-K2.6.
This indicates that failures are driven more by long-horizon state reasoning than by coding, planning, or domain-reasoning errors.
Among these errors, Cascade Error is the largest category, showing that incorrect intermediate states often propagate to later turns and affect downstream analyses.
State Management Error also contributes substantially, reflecting failures in selecting, updating, or restoring the correct analytical state.
By contrast, Context Memory Error occurs less frequently, suggesting that the main challenge is not merely retrieving prior information, but maintaining and applying the correct analytical state over time.

\subsection{Persistent State and Reset Effects}

\paragraph{Agentic Behavior Decreases over Long Trajectories.} 
We analyze how models' exploration behavior evolves as tasks progress.
As shown in Figure~\ref{fig:analysis}(d), the average number of agent steps per user turn decreases substantially, with the overall average dropping by \textbf{4.3} steps from early to late stages. 
This suggests that models explore the environment and establish code states in early turns, but increasingly rely on previously constructed states later.
As a result, they perform less exploration, verification, and iterative refinement, which may make early state errors harder to detect and correct, and contribute to downstream cascading failures, as shown in Section~\ref{subsec:error_analysis}.

\paragraph{Reset Experiment Setup.}
Motivated by the decrease in agent steps over long trajectories, we evaluate how resetting the code environment affects model performance.
During task execution, we reset the environment once at a task-specific turn, as detailed in Appendix~\ref{app:reset_experiment}.
We then compute \textbf{accuracy only on turns after the reset} and compare it with the corresponding persistent baseline, where the same turns are evaluated without resetting the environment.
To analyze how reset effects depend on the quality of the maintained state, we group tasks by their persistent accuracy on post-reset turns: Low (0--30\%), Medium (30--70\%), and High (70--100\%).

\paragraph{Reset Trades Off Recovery and State Continuity.}
Intuitively, resetting the code environment would hurt performance, as it removes accumulated variables, intermediate results, and other useful execution state.
However, Figure~\ref{fig:analysis}(e) shows a baseline-dependent effect: it slightly improves low- and medium-baseline cases, while substantially hurting high-baseline cases.
This suggests that reset can help when the persistent code state has drifted, since it clears potentially erroneous execution state and requires the model to reconstruct the needed analytical state from the interaction history.

Figure~\ref{fig:analysis}(f) shows a negative correlation between persistent post-reset accuracy and reset gain.
This pattern is consistent with reset helping degraded trajectories by reducing error propagation, while hurting strong trajectories by removing useful accumulated analytical state.
Overall, code-environment reset trades off recovery from corrupted states against preservation of state continuity.

\section{Related Work}
\label{sec:related}
\noindent \textbf{Data Analysis Benchmarks and Agents.}
Existing data science benchmarks have progressed from isolated coding tasks to agentic and interactive data analysis settings. 
Code-oriented benchmarks mainly evaluate standalone data science programming problems \citep{lai2023ds1000,huang2024dacodeagentdatascience}. 
Agentic benchmarks further require planning, code execution, tool use, and environment interaction within multi-step analytical workflows \citep{hu2024infiagent,zhang-etal-2024-benchmarking-data,gu-etal-2024-blade,jing2025dsbench,zhang2025datascibenchk,egg2025dabstep,chan2025mlebench,wang2025fdabench,weng2025unidatabench,ma2026aiagentsanswerdata}. 
Recent interactive benchmarks move closer to human data analysis by simulating analyst-agent collaboration and multi-round exploratory analysis \citep{dutta2025condabench,li2024tapilotcrossingbenchmarkingevolvingllms,li2025idabench,luo2026agentds}. 
In parallel, a growing line of data analysis agents automate end-to-end analytical pipelines via iterative reasoning and self-debugging \citep{rahman2025llmbaseddatascienceagents,zhu2025open,hong2024datainterpreter,you-etal-2025-datawiseagent,qiao2025scaling,nie2026dsgym,zhang2025deepanalyze,qiu2026rewardingscientificprocessprocesslevel}. 
However, existing benchmarks and systems still emphasize task completion or workflow automation, leaving long-horizon analytical state management underexplored.

\paragraph{Long-Horizon and Multi-Turn Agent Evaluation.}
Recent work has studied LLM agents in multi-turn and long-horizon settings \citep{liu2025agentbench,mialon2023gaia}, including dynamic user-agent interaction, tool use \citep{yao2024taubench,qin2023toolllm}, web or API operations, and extended workflow completion \citep{zhou2024webarena,drouin2024workarena,xie2024osworld}. 
These benchmarks reveal performance degradation under sustained interaction, long-range consistency requirements, and changing task contexts \citep{luo2025ultrahorizon}. 
However, they primarily target conversational tasks, web navigation, policy following, or general tool-use workflows rather than data analysis. 
\longds differs by grounding long-horizon interaction in stateful data analysis environments, where agents must maintain, revise, restore, and compose evolving analytical states across extended analysis sessions.
Extended discussion in Appendix~\ref{app:detailed-related}.

\section{Conclusion}
We introduce \longds, a benchmark for evaluating long-horizon agentic data analysis in stateful data-analysis environments, where agents must maintain, update, restore, and compose evolving analytical states across extended interactions.
Our results show that current proprietary and open-source models still struggle substantially in this setting, with performance degrading over longer trajectories and failures dominated by cascading and state-management errors.
By making these long-horizon state-management limitations explicit, \longds provides a challenging testbed for developing data-analysis agents that can more reliably manage analytical state over extended workflows.

\section*{Limitations}
While \longds provides a realistic benchmark for evaluating long-horizon agentic data analysis, several limitations remain. 

First, \longds is constructed from public Kaggle notebooks and datasets, which provide realistic analytical workflows but may not fully cover proprietary or production data-analysis scenarios. 
This also results in an imbalanced domain distribution, especially in Sports, where many candidate notebooks were filtered out due to large datasets or long-running computations.

Second, \longds emphasizes quantitatively verifiable questions to support reliable evaluation, and therefore only partially covers open-ended insight generation, visualization-heavy analysis, and presentation-oriented analytics. 

Third, \longds uses a semi-automated task construction pipeline with Codex-assisted generation and expert-guided refinement. 
Although manual review helps ensure task quality, the resulting tasks may still reflect biases from the source notebooks or the construction process.

\section*{Ethics Statement}
All datasets, notebooks, and the DSGYM framework used in this work are governed by their respective licenses, competition rules, and usage restrictions. We comply with all applicable terms.
We do not collect any new personal data in constructing the benchmark, and our benchmark tasks do not involve private or sensitive personal information.
Overall, we do not foresee any substantial ethical or societal concerns arising from this work.

\section*{Acknowledgements}
We would like to express our sincere gratitude to the anonymous reviewers for their thoughtful and constructive feedback. This work was supported by the New Generation Artificial Intelligence-National Science and Technology Major Project (2025ZD0122802), National Natural Science Foundation of China (No. 62576307, No. NSFCU23B2055, No. NSFCU19B2027), the Fundamental Research Funds for the Central Universities (226-2023-00138), the Yongjiang Talent Introduction Programme (2021A-156-G), and the Information Technology Center and State Key Lab of CAD\&CG, Zhejiang University.  
This work was also supported by Xinmiao Project of Zhejiang Provincial Applied Basic Research Program (2026XMZD064).
This work was supported by Ant Group and Zhejiang University - Ant Group Joint Laboratory of Knowledge Graph.

\bibliography{custom}




\clearpage

\appendix
\section{Details of Benchmark}

\subsection{State-Evolution Patterns}
\label{app:state-patterns}

\definecolor{StateBlue}{RGB}{226,240,255}
\definecolor{StateGreen}{RGB}{226,246,233}
\definecolor{StateYellow}{RGB}{255,246,214}
\definecolor{StatePurple}{RGB}{241,230,255}
\definecolor{StateRed}{RGB}{255,232,232}
\definecolor{StateTeal}{RGB}{224,247,247}

\makeatletter
\newcommand{\textsb}[1]{{\fontseries{sb}\selectfont #1}}
\makeatother

\newcommand{\keyhl}[2]{%
  \begingroup
  \setlength{\fboxsep}{1.2pt}%
  \colorbox{#1}{\textbf{#2}}%
  \endgroup
}

\begin{table*}[t!]
\centering
\small
\begin{tabular}{m{0.12\linewidth} m{0.40\linewidth} m{0.35\linewidth}}
\toprule
\textbf{Pattern} & \textbf{Definition} & \textbf{Example} \\
\midrule

Initial
& Establishes a \keyhl{StateBlue}{reusable analytical object}, such as a cohort, metric, rule, or intermediate result.
& Define \keyhl{StateBlue}{high-activity users} as those with at least 10 sessions. \\
\midrule

Inheritance
& Reuses the \keyhl{StateBlue}{most recent valid analytical state} without restating it.
& Using the \keyhl{StateBlue}{same user group}, compare retention across regions. \\
\midrule

Update
& Revises a previous definition, formula, filter, aggregation rule, or baseline, making the revision the \keyhl{StateBlue}{new default state}.
& Use 20 sessions as the \keyhl{StateBlue}{new cutoff} for high-activity users in the \keyhl{StateBlue}{following analysis}. \\
\midrule

Counterfactual
& Introduces a \keyhl{StateBlue}{temporary alternative assumption} for the \keyhl{StateBlue}{current turn only}.
& Recompute the result assuming a \keyhl{StateBlue}{5-session} cutoff instead. \\
\midrule

Rollback
& Answers under an \keyhl{StateBlue}{earlier anchored version} of the analysis instead of the most recent state.
& Revisit the \keyhl{StateBlue}{initial high-activity} definition and recompute the result. \\
\midrule

Composition
& Combines \keyhl{StateBlue}{two or more explicit state operations} beyond default inheritance.
& Use the \keyhl{StateBlue}{initial user group}, but evaluate it with the \keyhl{StateBlue}{revised retention metric}. \\
\bottomrule
\end{tabular}

\caption{
State-evolution patterns in \longds.
Short labels denote initial state construction, state inheritance, state update, counterfactual perturbation, rollback, and multi-state composition, respectively.
\protect\keyhl{StateBlue}{Blue} highlights mark the key state semantics in each definition and example.
Examples are illustrative.
}
\label{tab:state-patterns}
\end{table*}

Table~\ref{tab:state-patterns} summarizes the State-evolution patterns in \longds.
State inheritance is included for completeness, but it is treated as the default continuity assumption rather than a separate annotated category in benchmark statistics.

\subsection{Additional Benchmark Statistics}
\label{app:benchmark-statistics}

\begin{figure}[t]
\centering
\includegraphics[width=\linewidth]{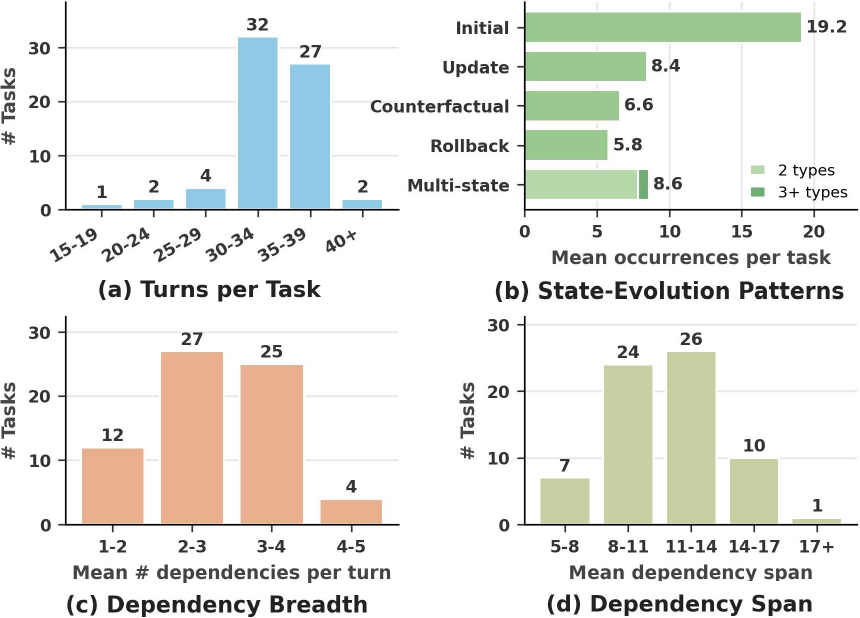}
\caption{
Benchmark statistics of \longds.
\textbf{(a) Turn Distribution}: Number of turns per task. 
\textbf{(b) State-Evolution Patterns}: Mean occurrences per task; Multi-state denotes turns annotated with two or more state types. 
\textbf{(c) Dependency Breadth}: Distribution of task-level mean number of direct prior-turn dependencies per turn.
\textbf{(d) Dependency Span}: Distribution of task-level mean farthest prior-turn dependency distance, in number of turns.
}
\label{fig:state-statistics}
\end{figure}

Figure~\ref{fig:state-statistics} shows the state-evolution patterns of \longds.
Table~\ref{tab:benchmark-macro-statistics} reports the task-level macro statistics of \longds. 

\begin{table}[t]
\centering
\small
\begin{tabular}{@{}lrrrrrr@{}}
\toprule
\textbf{Domain} & Edu. & Comm. & Soc. & Bus. & Geo. & Spo. \\
\midrule
\# Tasks & 8 & 16 & 10 & 12 & 19 & 3 \\
\midrule
\multicolumn{7}{@{}l@{}}{\textbf{Overall statistics} \quad \# Tasks = 68, \# Domains = 6} \\
\midrule
\multicolumn{3}{@{}l}{Turns / task} & \multicolumn{4}{r@{}}{32.7} \\
\multicolumn{3}{@{}l}{Initial / task} & \multicolumn{4}{r@{}}{19.2} \\
\multicolumn{3}{@{}l}{Update / task} & \multicolumn{4}{r@{}}{8.4} \\
\multicolumn{3}{@{}l}{Counterfactual / task} & \multicolumn{4}{r@{}}{6.6} \\
\multicolumn{3}{@{}l}{Rollback / task} & \multicolumn{4}{r@{}}{5.8} \\
\multicolumn{3}{@{}l}{Multi-state / task} & \multicolumn{4}{r@{}}{8.6} \\
\multicolumn{3}{@{}l}{Dependency breadth / turn} & \multicolumn{4}{r@{}}{2.85} \\
\multicolumn{3}{@{}l}{Dependency span / turn} & \multicolumn{4}{r@{}}{11.29} \\
\bottomrule
\end{tabular}
\caption{Benchmark scale and task-level macro statistics in \longds. Mean values are first computed within each task and then averaged across all 68 tasks.}
\label{tab:benchmark-macro-statistics}
\end{table}

\subsection{Task sources}
Table~\ref{tab:kaggle-sources} lists the Kaggle competitions, public datasets, and analysis notebooks used to construct \longds.
These sources were selected to cover diverse real-world analytical domains and to provide executable workflows with sufficient analytical depth for long-horizon task construction.
For each source, we report the corresponding task identifier, notebook title, and URL to make the benchmark provenance transparent and reproducible.

\section{Extended Related Work}
\label{app:detailed-related}

\paragraph{Surveys on data science agents.}
Recent surveys organize LLM-based data science agents by capability, autonomy, data-science lifecycle stage, and application setting \citep{rahman2025llmbaseddatascienceagents,zhu2026surveydataagentsemerging,chen2025largelanguagemodelbaseddata,Sun_2025}.
Work on LLM-based data preparation further covers data cleaning, transformation, and application-ready preparation pipelines \citep{zhou2026llmscleanmesssurvey}.

\paragraph{Data analysis and data science benchmarks.}
Prior benchmarks evaluate data-science coding, analytical reasoning, machine-learning experimentation, and agentic data workflows \citep{lai2023ds1000,huang2024dacodeagentdatascience,hu2024infiagent,zhang-etal-2024-benchmarking-data,gu-etal-2024-blade,jing2025dsbench,zhang2025datascibenchk,egg2025dabstep,huang2024mlagentbench,chan2025mlebench,luo-etal-2025-assistedds,wang2025fdabench,weng2025unidatabench,ma2026aiagentsanswerdata}.
Other recent benchmarks broaden evaluation to long-document analysis, heterogeneous document and table settings, spreadsheet environments, task-framing ambiguity, multimodal data science workflows, and cross-modal analysis \citep{li2026longdabenchmarkingllmagents,li2026dataclawautonomousdataagent,stoisser2026ambigdsbenchmarktaskframingambiguity,chi2026spreadsheetrladvancinglargelanguage,wu2025tablebench,yang2026aidabench,qi2026datacross,cao2024spider2}.

\paragraph{Interactive data analysis benchmarks.}
Interactive and conversational benchmarks evaluate agents under multi-round analytical interaction, guided analysis, or analyst-agent collaboration \citep{dutta2025condabench,li2024tapilotcrossingbenchmarkingevolvingllms,li2025idabench,luo2026agentds}.

\paragraph{Comparison with Existing Benchmarks.}

Table~\ref{tab:benchmark_comparison} compares \longds with recent benchmarks for agentic and interactive data analysis. 
\textsc{DABStep} and \textsc{LongDA} primarily evaluate single-turn data analysis tasks, while \textsc{AgentDS} focuses on comparing AI-only performance with human--AI collaborative performance. 
Although \textsc{ConDABench} and \textsc{IDA-Bench} support multi-turn interaction, their turns primarily serve to clarify underspecified requests or progressively provide task guidance.

In contrast, \longds evaluates the long-horizon management of evolving analytical states across multi-turn data analysis workflows. 
Each turn constitutes a concrete analytical request with its own reference answer and turn-level evaluation. 
This design enables direct assessment of whether an agent can correctly inherit, update, counterfactually modify, restore, and compose analytical states over extended interactions.

\paragraph{Data analysis agents and systems.}
LLM-based data analysis systems automate exploration, spreadsheet manipulation, visualization, notebook-centered analysis, data-science pipelines, and competition-style workflows \citep{ma-etal-2023-insightpilot,li2023sheetcopilot,yang2024matplotagent,guo2024dsagent,hong2024datainterpreter,you-etal-2025-datawiseagent,li2024autokaggle,zhang2025datacopilotbridgingbillionsdata,wang2025dataformulator2iterative}.
Recent systems and methods further emphasize scalable training, adaptive planning, verification, process supervision, and heterogeneous data analytics \citep{zhu2025open,qiao2025scaling,qiu2026rewardingscientificprocessprocesslevel,rewolinski2026sanitychecksagenticdata,zheng2026predict,nie2026dsgym,zhang2025deepanalyze,ou2025automindadaptiveknowledgeableagent,yang2025rdagentllmagentframeworkautonomous,nam2025mlestarmachinelearningengineering,nam2026dsstardatascienceagent,sun2025agenticdataagenticdataanalytics}.

\paragraph{Long-horizon and tool-use agent evaluation.}
General agent benchmarks study multi-turn interaction, tool use, web navigation, office workflows, API use, and long-horizon task completion \citep{liu2025agentbench,mialon2023gaia,yao2024taubench,qin2023toolllm,guo2025stabletoolbench,zhou2024webarena,dechezelles2025browsergym,drouin2024workarena,xie2024osworld,wang2025odysseybench,xu2025agentcompany,luo2025ultrahorizon,abhyankar2026osworldhuman,jiang2026agentlab,zhang2026mmskillsmultimodalskillsgeneral}.
Broader work on tool-using, reflective, memory-augmented, and multi-agent LLM systems provides additional context for agent design and evaluation \citep{schick2023toolformer,gao2023pal,wu2023autogen,qiao2024taskweaver,shinn2023reflexion,madaan2023selfrefine,wang2023voyager,packer2024memgpt,park2023generativeagents,sun2026CHIMERA,zhong2023memorybank}.

\section{Details of Experiment}
\label{app:experiment-details}

\subsection{Experimental Setup}
The data analysis agent is instructed via a unified, structured system prompt designed to regulate its tool-use and reasoning format within the DSGYM framework. The system prompt is provided in Appendix~\ref{prompt:system_prompt}.

To establish a controlled evaluation environment, we standardize the generation parameters across all baseline models to the furthest extent permitted by their respective APIs. For GPT-5.4, Gemini-3.1-Pro, Claude-4.6-Sonnet, and DeepSeek-V4-Pro, the temperature is strictly set to $0.0$. For Kimi-K2.6, the temperature is maintained at $1.0$ because the official API implementation restricts temperature modification. Across all evaluated models, the maximum number of output tokens is capped at $8,192$ per interaction turn to accommodate extensive multi-step analytical outputs.

\subsection{Codex Results on a Sampled Subset}
\label{app:codex_results}

Evaluating Codex requires a manual turn-by-turn interaction protocol: each task turn must be provided to Codex sequentially, and each turn-level answer must be manually collected before proceeding to the next turn.
Due to this operational cost, we evaluate \textit{Codex} (GPT-5.4, with high reasoning effort) on a domain-stratified sampled subset of \longds, selecting two tasks from each of the six domains.

For comparison, Table~\ref{tab:codex-sampled-results} reports the performance of \textit{Codex} and the other evaluated models on the same sampled subset.
The results should therefore be interpreted as a complementary sampled-subset study rather than a replacement for the full benchmark results in Table~\ref{tab:main_results}.
On this sampled subset, \textit{Codex} achieves the highest average score of 65.55 and shows strong performance across several domains.  
These results suggest that a stronger code-centric agent can improve performance on some LongDS tasks, while long-horizon analytical state management remains challenging.

\begin{table*}[t]
\centering
\small
\resizebox{\textwidth}{!}{%
\begin{tabular}{@{}lcccccc|c@{}}
\toprule
\textbf{Model} & \textbf{Community} & \textbf{Education} & \textbf{Business} & \textbf{Geoscience} & \textbf{Social Good} & \textbf{Sports} & \textbf{Average} \\
\midrule
Codex & \cellcolor{cyan!18}\textbf{56.67} & \cellcolor{cyan!18}\textbf{86.67} & 67.10 & 68.50 & \cellcolor{cyan!18}\textbf{80.00} & \cellcolor{cyan!18}\textbf{34.34} & \cellcolor{cyan!18}\textbf{65.55} \\
GPT-5.4 & 41.67 & 70.00 & \cellcolor{cyan!8}75.65 & 68.58 & 71.67 & 13.40 & 56.98 \\
Claude-4.6 & \cellcolor{cyan!8}55.00 & 63.00 & \cellcolor{cyan!18}\textbf{77.97} & 70.59 & 71.67 & 15.36 & 59.26 \\
Gemini-3.1 & \cellcolor{cyan!8}55.00 & \cellcolor{cyan!8}75.00 & 70.95 & \cellcolor{cyan!18}\textbf{82.12} & 73.34 & 15.63 & \cellcolor{cyan!8}61.17 \\
Kimi-K2.6 & 41.67 & 73.00 & 62.61 & \cellcolor{cyan!8}77.56 & 73.34 & \cellcolor{cyan!8}19.51 & 57.78 \\
DeepSeek-V4 & 38.34 & 69.34 & 66.45 & 19.04 & \cellcolor{cyan!8}76.67 & 11.84 & 46.61 \\
\bottomrule
\end{tabular}
}
\caption{
Codex results on a domain-stratified sampled subset of \longds.
We sample two tasks from each domain and evaluate Codex using a manual turn-by-turn protocol.
All scores are accuracies (\%) computed on the same sampled subset.
\protect\colorbox{cyan!18}{\textbf{Best}} and
\protect\colorbox{cyan!8}{second-best} scores are highlighted.
}
\label{tab:codex-sampled-results}
\end{table*}

\subsection{Statistical Information Details}
\label{app:statistical}

\begin{table*}[t]
\centering
\small
\setlength{\tabcolsep}{4pt}
\begin{tabular}{lccccccc}
\toprule
\textbf{Model} 
& \textbf{Avg (n=68)}
& \textbf{Edu (n=8)}
& \textbf{Comm (n=16)}
& \textbf{Soc (n=10)}
& \textbf{Bus (n=12)}
& \textbf{Geo (n=19)}
& \textbf{Spo (n=3)} \\
\midrule
GPT
& 43.5 {\scriptsize [36.1, 50.9]}
& 77.9 {\scriptsize [69.3, 86.3]}
& 65.3 {\scriptsize [51.5, 78.4]}
& 36.8 {\scriptsize [25.3, 50.5]}
& 28.4 {\scriptsize [22.3, 34.6]}
& 28.9 {\scriptsize [15.9, 43.3]}
& 10.5 {\scriptsize [4.8, 14.3]} \\

Claude
& 41.6 {\scriptsize [34.8, 48.4]}
& 77.3 {\scriptsize [63.8, 89.6]}
& 54.6 {\scriptsize [40.8, 67.9]}
& 36.1 {\scriptsize [22.8, 50.9]}
& 25.5 {\scriptsize [17.3, 33.8]}
& 31.9 {\scriptsize [20.5, 44.7]}
& 19.8 {\scriptsize [5.7, 28.6]} \\

Gemini
& 48.5 {\scriptsize [41.8, 55.0]}
& 58.0 {\scriptsize [39.4, 75.6]}
& 69.5 {\scriptsize [58.7, 78.5]}
& 41.7 {\scriptsize [28.0, 55.8]}
& 33.6 {\scriptsize [22.9, 43.9]}
& 42.2 {\scriptsize [29.3, 55.3]}
& 31.9 {\scriptsize [0.0, 64.3]} \\

Kimi
& 39.7 {\scriptsize [33.1, 46.3]}
& 65.0 {\scriptsize [44.3, 81.7]}
& 60.6 {\scriptsize [50.3, 70.1]}
& 31.3 {\scriptsize [16.8, 48.2]}
& 21.0 {\scriptsize [16.2, 26.1]}
& 28.8 {\scriptsize [17.6, 41.5]}
& 32.9 {\scriptsize [17.1, 59.5]} \\

DeepSeek
& 32.0 {\scriptsize [25.7, 38.4]}
& 61.4 {\scriptsize [42.1, 77.5]}
& 49.5 {\scriptsize [36.9, 61.8]}
& 32.4 {\scriptsize [19.3, 48.1]}
& 17.1 {\scriptsize [10.9, 25.2]}
& 16.6 {\scriptsize [9.5, 25.6]}
& 15.8 {\scriptsize [9.4, 23.8]} \\
\bottomrule
\end{tabular}

\caption{Per-domain performance with 95\% bootstrap confidence intervals.
Edu, Comm, Soc, Bus, Geo, and Spo denote Education, Community, Social Good,
Business, Geoscience, and Sports, respectively.}
\label{tab:domain_ci}
\end{table*}
\begin{table*}[t]
\centering
\small
\setlength{\tabcolsep}{6pt}
\begin{tabular}{lcccc}
\toprule
\textbf{Model}
& \textbf{Initial}
& \textbf{Update}
& \textbf{Counterfactual}
& \textbf{Rollback} \\
\midrule
GPT
& 46.8 [38.8, 54.5]
& 35.0 [28.7, 41.9]
& 28.9 [21.3, 37.2]
& 20.8 [13.5, 29.0] \\

Claude
& 44.2 [37.5, 50.9]
& 31.0 [25.0, 37.4]
& 31.5 [24.6, 39.3]
& 21.1 [15.1, 27.8] \\

Gemini
& 52.4 [45.5, 58.8]
& 41.1 [34.0, 48.2]
& 36.7 [28.3, 45.3]
& 33.0 [24.3, 42.4] \\

Kimi
& 43.0 [36.2, 49.6]
& 30.1 [24.1, 37.0]
& 26.4 [20.0, 33.4]
& 19.3 [13.4, 25.9] \\

DeepSeek
& 35.7 [29.2, 42.1]
& 23.9 [18.7, 29.8]
& 19.2 [13.5, 25.7]
& 16.8 [10.7, 23.9] \\
\midrule
Overall
& 44.4 [38.1, 50.6]
& 32.2 [27.0, 38.0]
& 28.5 [22.7, 34.9]
& 22.2 [16.5, 28.8] \\
\bottomrule
\end{tabular}
\caption{Performance across state-evolution patterns with 95\% bootstrap confidence intervals.}
\label{tab:pattern_ci}
\end{table*}
To assess the statistical reliability of our results, we report bootstrap 95\% confidence intervals for both \textbf{per-domain} and \textbf{per-pattern} performance, together with \textbf{three-run evaluations} of two representative models.

For the \textbf{domain-level results}, we perform task-level bootstrap resampling for each model--domain pair and recompute the average score for each bootstrap sample.
This accounts for variability across tasks within each domain. 
The resulting confidence intervals are reported in Table~\ref{tab:domain_ci}. 
As expected, domains with fewer tasks exhibit wider confidence intervals.

For the \textbf{state-evolution pattern results}, pattern labels are assigned at the turn level, while multiple turns may originate from the same task and are therefore not independent. We consequently adopt a task-cluster bootstrap: tasks are resampled with replacement, and pattern-level accuracy is recomputed over all corresponding turns from the sampled tasks. This preserves within-task dependencies and avoids treating individual turns as independent observations. The resulting 95\% confidence intervals are reported in Table~\ref{tab:pattern_ci}.

\begin{table*}[t]
\centering
\small
\setlength{\tabcolsep}{5pt}
\begin{tabular}{lccccccc}
\toprule
\textbf{Model}
& \textbf{Edu}
& \textbf{Comm}
& \textbf{Soc}
& \textbf{Bus}
& \textbf{Geo}
& \textbf{Spo}
& \textbf{Avg} \\
\midrule

GPT-5.4
& 73.98 {\scriptsize $\pm$ 4.96}
& 59.07 {\scriptsize $\pm$ 4.43}
& 35.74 {\scriptsize $\pm$ 3.09}
& 31.01 {\scriptsize $\pm$ 3.60}
& 28.52 {\scriptsize $\pm$ 1.54}
& 24.38 {\scriptsize $\pm$ 9.98}
& 42.37 {\scriptsize $\pm$ 0.81} \\

DeepSeek-V4-Pro
& 64.09 {\scriptsize $\pm$ 1.97}
& 49.63 {\scriptsize $\pm$ 1.29}
& 29.95 {\scriptsize $\pm$ 2.43}
& 15.21 {\scriptsize $\pm$ 1.51}
& 22.26 {\scriptsize $\pm$ 4.61}
& 17.40 {\scriptsize $\pm$ 5.42}
& 33.29 {\scriptsize $\pm$ 0.95} \\
\bottomrule
\end{tabular}

\caption{Three-run evaluation results reported as mean $\pm$ standard deviation.
Edu, Comm, Soc, Bus, Geo, and Spo denote Education, Community, Social Good,
Business, Geoscience, and Sports, respectively.}
\label{tab:multi_run}
\end{table*}
For the \textbf{multi-run evaluation}, we conduct three independent runs for two representative models, GPT-5.4 and DeepSeek-V4-Pro, and report the mean and standard deviation across runs.
As shown in Table~\ref{tab:multi_run}, the overall scores exhibit relatively small run-to-run variation, with standard deviations of 0.81 for GPT-5.4 and 0.95 for DeepSeek-V4-Pro.
Variation is larger for some individual domains, particularly those with fewer tasks.

\subsection{Training-Data Contamination Analysis}
\label{app:contamination}

Since \longds is constructed from public Kaggle notebooks, we consider potential training-data contamination from two perspectives: mitigation during benchmark construction and empirical contamination probes.

\paragraph{Mitigation in Benchmark Construction.}
In \longds, source notebooks serve only as references for the overall analytical workflow, while benchmark questions do not directly reuse the original analyses or answers.
During task construction, we introduce new state-evolution operations, including updates, rollbacks, counterfactual perturbations, and compositions, yielding analytical trajectories that are absent from the source notebooks.
In addition, \longds requires structured, multi-field numerical answers, such as ranked lists of entities with their corresponding scores, rather than directly reusing notebook conclusions.
These design choices reduce the extent to which benchmark tasks can be solved by recalling the original notebook trajectories or conclusions.

\paragraph{N-gram Containment.}
We assess potential verbatim overlap by comparing each complete benchmark task, including its questions, solution code, and answers, with the corresponding source notebook using directed n-gram containment.
As shown in Table~\ref{tab:ngram_contamination}, containment decreases as $n$ increases, from 2.32\% for 4-grams to 0.94\% for 13-grams.
The low long-span containment rates suggest limited verbatim overlap between \longds tasks and their source notebooks.

\begin{table}[t]
\centering
\small
\begin{tabular}{lc}
\toprule
\textbf{Metric} & \textbf{Mean $\pm$ Std} \\
\midrule
4-gram containment  & 2.32 $\pm$ 4.14\% \\
6-gram containment  & 1.63 $\pm$ 3.60\% \\
8-gram containment  & 1.32 $\pm$ 3.28\% \\
10-gram containment & 1.13 $\pm$ 3.03\% \\
13-gram containment & 0.94 $\pm$ 2.73\% \\
\bottomrule
\end{tabular}
\caption{Directed n-gram containment between \longds tasks and their corresponding source notebooks.}
\label{tab:ngram_contamination}
\end{table}

\paragraph{No-Data Memorization Probe.}
We further evaluate whether models can answer benchmark questions without access to the underlying data files.
In this setting, models receive only the task context and question.
As shown in Table~\ref{tab:no_data_probe}, both GPT-5.4 and DeepSeek-V4-Pro achieve near-zero scores, with averages of 1.72 and 1.24, respectively.
These results suggest that direct answer memorization is unlikely to account for the observed benchmark performance.

\begin{table}[t]
\centering
\small
\begin{tabular}{lc}
\toprule
\textbf{Model} & \textbf{Avg. Score} \\
\midrule
DeepSeek-V4-Pro & 1.24 $\pm$ 0.16 \\
GPT-5.4         & 1.72 $\pm$ 0.16 \\
\bottomrule
\end{tabular}
\caption{No-data memorization probe. Models are evaluated without access to the underlying data files. Results are reported as mean $\pm$ standard deviation.}
\label{tab:no_data_probe}
\end{table}

\subsection{Error Analysis Details}
\label{app:error_analysis}

\paragraph{Agent-as-Judge Annotation.}
We use \textit{Codex} as an agent-as-judge to assist the annotation of incorrect turns with the prompt in Appendix~\ref{prompt:error_prompt}. 
Specifically, we run \textit{Codex} with GPT-5.5~\citep{OpenAI-GPT55} and x-high reasoning effort to examine the task context, reference answer, model response, and execution trace, and assign each incorrect turn to one of the predefined error categories.
For each evaluated task, we provide \textit{Codex} with four files: \texttt{code.py}, \texttt{ground\_code.py}, \texttt{results\_eval.json}, and \texttt{task.ipynb}. 
Together, these files contain the agent-generated code, reference solution code, turn-level questions and answers, evaluation outcomes, agent trajectories, and reference solution logic.

We sample six tasks from each domain, except for Sports where only three tasks are available, and annotate the results of all five evaluated models on each sampled task.
In total, Codex produces error annotations for 3,207 incorrect turns, which form the annotated error pool used for subsequent analysis and human validation.
The results are shown in Figure~\ref{fig:analysis}(c)

\paragraph{Human Validation of Error Annotations.}
We conduct two complementary human validation studies to assess the reliability of our error annotations.

First, in the \textbf{blind relabeling study}, we sample 200 error cases.
Annotators are shown the task information and model outputs but not the original error labels, and are asked to independently assign the primary error type according to our taxonomy.
For cases where annotators are uncertain, we conduct a follow-up discussion to clarify the applicable taxonomy criteria and finalize the annotation.
This setting evaluates whether the taxonomy can be applied consistently without label hints.

Second, in the \textbf{reference verification study}, we sample another 200 error cases, aiming to cover different error categories as well as boundary cases that are difficult to distinguish.
Annotators are shown the reference primary error type and its supporting evidence, and are asked to judge whether the reference label is appropriate; if they disagree, they provide a corrected primary error type.

In both studies, the 200 cases are approximately evenly assigned to three annotators, with each annotator reviewing about one third of the cases.
Annotators are provided with the error taxonomy and annotation guidelines in Appendix~\ref{prompt:error_prompt} before labeling.
We compute agreement, Cohen's $\kappa$, and macro-F1 by comparing human annotations with the corresponding Codex-generated primary error labels.

As shown in Table~\ref{tab:error-human-validation}, the blind relabeling study achieves 81.50\% overall agreement with a Cohen's $\kappa$ of 0.7535, indicating substantial agreement even when annotators do not see the original labels. 
The reference verification study yields higher agreement, with 89.00\% agreement, a Cohen's $\kappa$ of 0.8715, and a macro-F1 of 0.8898. 
These results suggest that the error taxonomy is generally reproducible under human annotation, while the lower macro-F1 in blind relabeling indicates that some boundary cases remain difficult to distinguish.

\begin{table*}[t]
\centering
\small
\begin{tabular}{llrrrr}
\toprule
Study & Annotator & Cases & Agreement & Cohen's $\kappa$ & Macro-F1 \\
\midrule
Blind relabeling & Human 1 & 67 & 82.09 & 0.7587 & 0.6403 \\
Blind relabeling & Human 2 & 67 & 77.61 & 0.7094 & 0.5306 \\
Blind relabeling & Human 3 & 66 & 84.85 & 0.7948 & 0.6725 \\
Blind relabeling & Overall & 200 & 81.50 & 0.7535 & 0.6177 \\
\midrule
Reference verification & Human 1 & 67 & 88.06 & 0.8605 & 0.8791 \\
Reference verification & Human 2 & 67 & 94.03 & 0.9302 & 0.9402 \\
Reference verification & Human 3 & 66 & 84.85 & 0.8228 & 0.8540 \\
Reference verification & Overall & 200 & 89.00 & 0.8715 & 0.8898 \\
\bottomrule
\end{tabular}
\caption{
Human validation results for error annotation reliability.
Agreement is reported as percentage accuracy, while Cohen's $\kappa$ and macro-F1 are computed against the original or reference primary error type.
}
\label{tab:error-human-validation}
\end{table*}

\subsection{Dependency Span Analysis}
\label{app:additional_analysis}

\begin{figure}[t]
    \centering
    \includegraphics[width=\linewidth]{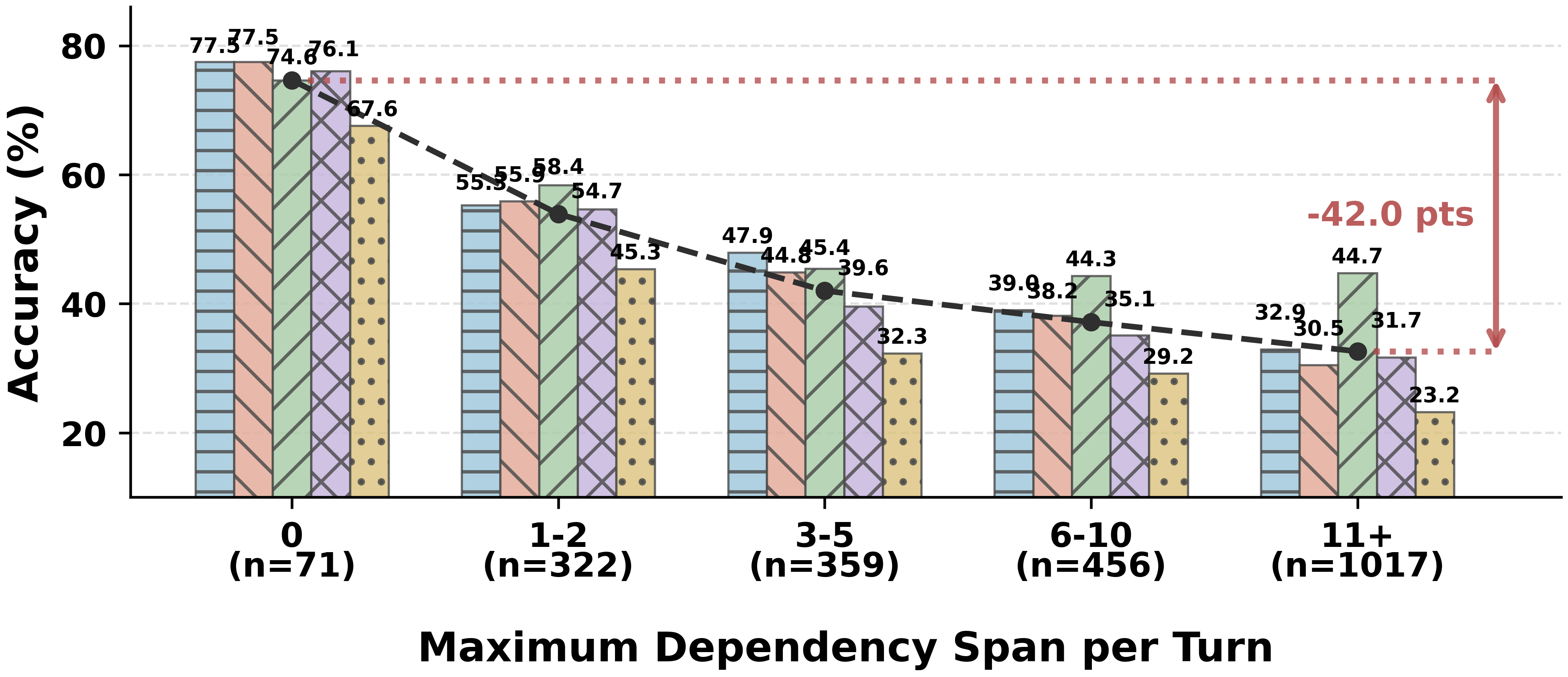}
    \caption{
    \textbf{Accuracy by dependency span.}
    Model accuracy decreases as the farthest dependency span becomes longer, indicating that agents struggle when the required analytical state must be recovered from more distant prior turns. 
    $n$ denotes the number of turns in each group.
    }
    \label{fig:appendix-span}
\end{figure}

Figure~\ref{fig:appendix-span} complements the dependency-breadth analysis in the main text. 
While dependency breadth measures how many prior turns a request directly depends on, dependency span measures the maximum distance to the farthest depended prior turn. 
Accuracy declines as dependency span increases, showing that long-range analytical dependencies further challenge agents' ability to recover and apply the correct analytical state.

\subsection{Reset Experiment}
\label{app:reset_experiment}

We conduct the reset experiment with GPT-5.4. 
We exclude five tasks for which the persistent baseline is either entirely incorrect or answers only one turn correctly, since such cases provide too little reliable state to analyze reset effects.
For each remaining task, we run the agent in the standard persistent setting and then reset the executable environment once at a task-specific turn.
The reset clears the accumulated code state, variables, and intermediate results in the execution environment, while the interaction history remains available to the agent.

To choose the reset point, we use a simple task-specific heuristic over four predefined candidate turns, 2, 4, 6, and 15. 
The candidate set covers early and middle stages of the trajectory while leaving enough turns for post-reset evaluation. 
For each task, we select the candidate whose remaining-turn ratio is closest to half of the task's persistent baseline accuracy. 
This heuristic is used only to define a consistent reset location for diagnostic comparison; all reported reset effects are computed on the same post-reset turns for both the reset run and the persistent baseline.

\section{Details of Evaluation}
\label{app:evaluation-details}

\begin{table}[t]
\centering
\small
\begin{tabular}{lrrrr}
\toprule
Annotator & Cases & Agreement & Cohen's $\kappa$ & Macro-F1 \\
\midrule
Human 1 & 150 & 92.00 & 0.8403 & 0.9199 \\
Human 2 & 150 & 94.67 & 0.8933 & 0.9466 \\
Human 3 & 150 & 92.67 & 0.8533 & 0.9266 \\
\midrule
Overall & 450 & 93.11 & 0.8623 & 0.9311 \\
\bottomrule
\end{tabular}
\caption{
Human audit results for LLM-as-judge evaluation.
Agreement is computed between blind human judgments and the original LLM-as-judge scores.
}
\label{tab:human_check_judge}
\end{table}

\begin{table}[t]
\centering
\small
\begin{tabular}{lrr}
\toprule
\multirow{2}{*}{LLM-as-judge score} & \multicolumn{2}{c}{Human score} \\
\cmidrule(lr){2-3}
 & 0 & 1 \\
\midrule
0 & 212 & 9 \\
1 & 22 & 207 \\
\bottomrule
\end{tabular}
\caption{
Confusion matrix between blind human judgments and LLM-as-judge scores.
}
\label{tab:judge_human_confusion}
\end{table}

\subsection{Evaluation Prompt}
The complete prompt template utilized for the automated LLM-as-a-judge evaluation is presented in \ref{prompt:eval_prompt}.

\subsection{Human Validation of the LLM Evaluator}
\label{app:judge-validation}
To validate the reliability of the automated LLM-as-judge evaluation, we conduct a blind human audit on 450 evaluated samples.
The sample is balanced across both domains and models, with 75 cases from each of the six domains and 90 cases from each of the five evaluated models.
We also balance the original judge labels, including 225 cases judged correct and 225 cases judged incorrect by the automated evaluator.

The 450 cases are split into three groups of 150 cases, each reviewed by one of three annotators.
Annotators are shown only the task question, ground-truth answer, and model response, without access to the original judge score or judge rationale.
They are given the same evaluation instructions as the LLM-as-judge prompt in Appendix~\ref{prompt:eval_prompt} and independently make a binary decision on whether the model response correctly answers the question.

We then compare the human judgments with the original LLM-as-judge scores and report agreement, Cohen's $\kappa$, and macro-F1.
As shown in Table~\ref{tab:human_check_judge}, the blind human audit shows high agreement between human judgments and the LLM-as-judge scores. 
Overall agreement reaches 93.11\%, with a Cohen's $\kappa$ of 0.8623 and a macro-F1 of 0.9311. 
These results suggest that the automated evaluator is reliable for turn-level answer correctness evaluation.

Table~\ref{tab:judge_human_confusion} further shows the confusion matrix between blind human judgments and LLM-as-judge scores.
Among 450 audited cases, 419 receive the same label from humans and the automated judge, while 31 cases differ.
The automated judge marks 22 human-incorrect answers as correct and 9 human-correct answers as incorrect.

\definecolor{StateBlue}{RGB}{226,240,255}
\definecolor{StateGreen}{RGB}{226,246,233}
\definecolor{StateYellow}{RGB}{255,246,214}
\definecolor{StatePurple}{RGB}{241,230,255}
\definecolor{StateRed}{RGB}{255,232,232}
\definecolor{StateTeal}{RGB}{224,247,247}

\makeatletter
\setlength{\@fptop}{0pt}
\setlength{\@fpsep}{8pt plus 2pt}
\setlength{\@fpbot}{0pt plus 1fil}
\makeatother

\newcolumntype{K}[1]{>{\centering\arraybackslash}m{#1}}
\newcolumntype{U}[1]{>{\RaggedRight\arraybackslash}m{#1}}
\newcommand{\SourceMultirow}[3][0pt]{\multirow[c]{#2}{0.24\textwidth}[#1]{\centering #3}}

\begin{table*}[p]
\centering
\scriptsize
\hypersetup{urlcolor=black}
\setlength{\tabcolsep}{2pt}
\renewcommand{\arraystretch}{1.03}
\caption{Kaggle Datasets and Notebooks in the Business Domain}
\label{tab:kaggle-sources}
\label{tab:kaggle-sources-business}
\begin{tabular}{@{}K{0.24\textwidth}K{0.08\textwidth}K{0.31\textwidth}U{0.33\textwidth}@{}}
\toprule
\textbf{Competition/Dataset} & \textbf{Task ID} & \textbf{Notebook Title} & \textbf{URL} \\
\midrule
goodbooks-10k & task1 & Netflix Vs Books-Recommender, Analysis, EDA & \url{https://www.kaggle.com/code/niharika41298/netflix-vs-books-recommender-analysis-eda} \\
\midrule
Goodreads-books & task1 & Book Recommendation Engine & \url{https://www.kaggle.com/code/snanilim/book-recommendation-engine} \\
\midrule
\SourceMultirow[-0.35\baselineskip]{2}{My Uber Drives~\citep{zeeshan-ul-hassan_usmani_2017}} & task1 & UBER Rides Dataset 2016 ANALYSIS & \url{https://www.kaggle.com/code/suiyue/uber-rides-dataset-2016-analysis} \\*
\cline{2-4}
  & task2 & Uber\_ride Analysis & \url{https://www.kaggle.com/code/saurav9786/uber-ride-analysis} \\
\midrule
\SourceMultirow[-1.05\baselineskip]{4}{Netflix Movies and TV Shows} & task1 & Netflix Visualizations, Recommendation, EDA & \url{https://www.kaggle.com/code/niharika41298/netflix-visualizations-recommendation-eda} \\*
\cline{2-4}
  & task2 & storytelling with Data - Netflix ver. & \url{https://www.kaggle.com/code/subinium/storytelling-with-data-netflix-ver} \\*
\cline{2-4}
  & task3 & NETFLIX CONSUMPTION ANALYSIS & \url{https://www.kaggle.com/code/sahib12/netflix-consumption-analysis/notebook} \\*
\cline{2-4}
  & task4 & Netflix Data Visualization & \url{https://www.kaggle.com/code/joshuaswords/netflix-data-visualization} \\
\midrule
\SourceMultirow[-0.7\baselineskip]{3}{NYC Restaurants Data - Food Ordering and Delivery} & task1 & NYC Restaurant Food Order \& Delivery Detailed EDA & \url{https://www.kaggle.com/code/ahsan81/nyc-restaurant-food-order-delivery-detailed-eda} \\*
\cline{2-4}
  & task2 & Exploratory Data Analysis - NYC FoodHub & \url{https://www.kaggle.com/code/lilyhyseni/exploratory-data-analysis-nyc-foodhub} \\*
\cline{2-4}
  & task3 & Delivery Time - EDA, Grouping and ML (32\%) & \url{https://www.kaggle.com/code/raphaelmarconato/delivery-time-eda-grouping-and-ml-32} \\
\midrule
Transaction Data for fraud analysis & task1 & Transaction Data for fraud analysis & \url{https://www.kaggle.com/code/neamulislamfahim/transaction-data-for-fraud-analysis} \\
\bottomrule
\end{tabular}
\end{table*}

\begin{table*}[p]
\centering
\scriptsize
\hypersetup{urlcolor=black}
\setlength{\tabcolsep}{2pt}
\renewcommand{\arraystretch}{1.03}
\caption{Kaggle Datasets and Notebooks in the Education Domain}
\label{tab:kaggle-sources-education}
\begin{tabular}{@{}K{0.24\textwidth}K{0.08\textwidth}K{0.31\textwidth}U{0.33\textwidth}@{}}
\toprule
\textbf{Competition/Dataset} & \textbf{Task ID} & \textbf{Notebook Title} & \textbf{URL} \\
\midrule
\SourceMultirow[-0.35\baselineskip]{2}{BI intro to data cleaning eda and machine learning} & task1 & BI Data Cleaning, EDA and Predictive Modeling & \url{https://www.kaggle.com/code/lukhilaksh/bi-data-cleaning-eda-and-predictive-modeling} \\*
\cline{2-4}
  & task2 & notebook524401d43e & \url{https://www.kaggle.com/code/walekhwatlphilip/notebook524401d43e} \\
\midrule
\SourceMultirow[-0.7\baselineskip]{3}{LearnPlatform COVID-19 Impact on Digital Learning~\citep{learnplatform-covid19-impact-on-digital-learning}} & task1 & Maslow Before Bloom & \url{https://www.kaggle.com/code/iamleonie/maslow-before-bloom/input} \\*
\cline{2-4}
  & task2 & Learning in Cyberspace: a Story of Pandemic Times & \url{https://www.kaggle.com/code/mauromauro/learning-in-cyberspace-a-story-of-pandemic-times/notebook\#8.-Wrap-up} \\*
\cline{2-4}
  & task3 & Digital Learning During Pandemic-Contest Winner & \url{https://www.kaggle.com/code/charliezimmerman/digital-learning-during-pandemic-contest-winner/notebook?scriptVersionId=87449728} \\
\midrule
\SourceMultirow[-0.7\baselineskip]{3}{World University Rankings} & task1 & World University Rankings Advanced Analysis & \url{https://www.kaggle.com/code/gpreda/world-university-rankings-advanced-analysis\#conclusions} \\*
\cline{2-4}
  & task2 & Which universities do good science? & \url{https://www.kaggle.com/code/pozdniakov/which-universities-do-good-science} \\*
\cline{2-4}
  & task3 & MSU vs Top-7 & \url{https://www.kaggle.com/code/ospanoff/msu-vs-top-7/notebook} \\
\bottomrule
\end{tabular}
\end{table*}

\begin{table*}[p]
\centering
\scriptsize
\hypersetup{urlcolor=black}
\setlength{\tabcolsep}{2pt}
\renewcommand{\arraystretch}{1.03}
\caption{Kaggle Datasets and Notebooks in the Geoscience Domain}
\label{tab:kaggle-sources-geoscience}
\begin{tabular}{@{}K{0.24\textwidth}K{0.08\textwidth}K{0.31\textwidth}U{0.33\textwidth}@{}}
\toprule
\textbf{Competition/Dataset} & \textbf{Task ID} & \textbf{Notebook Title} & \textbf{URL} \\
\midrule
\SourceMultirow[-0.7\baselineskip]{3}{Acea Smart Water Analytics~\citep{acea-water-prediction}} & task1 & Intro to Time Series Forecasting & \url{https://www.kaggle.com/code/iamleonie/intro-to-time-series-forecasting} \\*
\cline{2-4}
  & task2 & How virtual trees can save real water in Italy? & \url{https://www.kaggle.com/code/michau96/how-virtual-trees-can-save-real-water-in-italy} \\*
\cline{2-4}
  & task3 & Acea Smart Water: Full EDA \& Prediction & \url{https://www.kaggle.com/code/maksymshkliarevskyi/acea-smart-water-full-eda-prediction} \\
\midrule
\SourceMultirow[-1.05\baselineskip]{4}{CDP - Unlocking Climate Solutions~\citep{cdp-unlocking-climate-solutions}} & task1 & KPIs for measuring Climate Action and Inequality & \url{https://www.kaggle.com/code/mannmann2/kpis-for-measuring-climate-action-and-inequality} \\*
\cline{2-4}
  & task2 & CDP Challenge: Climate Adaptation Index & \url{https://www.kaggle.com/code/shabou/cdp-challenge-climate-adaptation-index/report} \\*
\cline{2-4}
  & task3 & CDP: A Path to Efficient and Sustainable Growth & \url{https://www.kaggle.com/code/katemelianova/cdp-a-path-to-efficient-and-sustainable-growth} \\*
\cline{2-4}
  & task4 & Impact Potential Analysis of Water-Use Efficiency & \url{https://www.kaggle.com/code/iamleonie/impact-potential-analysis-of-water-use-efficiency} \\
\midrule
\SourceMultirow[-0.35\baselineskip]{2}{DS4G - Environmental Insights Explorer~\citep{ds4g-environmental-insights-explorer}} & task1 & DS4G: An analytical approach to NO2 emissions & \url{https://www.kaggle.com/code/chrisarderne/ds4g-an-analytical-approach-to-no2-emissions/notebook} \\*
\cline{2-4}
  & task2 & DS4G: Spatial Panel Data Modeling & \url{https://www.kaggle.com/code/katemelianova/ds4g-spatial-panel-data-modeling} \\
\midrule
\SourceMultirow[-0.7\baselineskip]{3}{Global Data on Sustainable Energy (2000-2020)~\citep{ansh_tanwar_2023}} & task1 & Starter Notebook: Global Sustainable Energy & \url{https://www.kaggle.com/code/anshtanwar/starter-notebook-global-sustainable-energy} \\*
\cline{2-4}
  & task2 & EcoOpt & \url{https://www.kaggle.com/code/ahmadihossein/ecoopt} \\*
\cline{2-4}
  & task3 & EDA |CO2 emission Data | Visualization & \url{https://www.kaggle.com/code/abdallhaosman/eda-co2-emission-data-visualization} \\
\midrule
Marmara Region Earthquakes (Apr 23--24, 2025) & task1 & Istanbul Quake Watch: Forecasting the Megaquake & \url{https://www.kaggle.com/code/pinuto/istanbul-quake-watch-forecasting-the-megaquake} \\
\midrule
\SourceMultirow[-0.7\baselineskip]{3}{Excellence in Research Award (Phase II)~\citep{phase-ii-widsdatathon2022}} & task1 & WIDS II 2022: EDA & \url{https://www.kaggle.com/code/sytuannguyen/wids-ii-2022-eda} \\*
\cline{2-4}
  & task2 & Learning with Our Vulnerability: Covid-19 & \url{https://www.kaggle.com/code/mpwolke/learning-with-our-vulnerability-covid-19/notebook} \\*
\cline{2-4}
  & task3 & How well (or not) we live - Health Rankings WiDS & \url{https://www.kaggle.com/code/mpwolke/how-well-or-not-we-live-health-rankings-wids} \\
\midrule
\SourceMultirow[-0.7\baselineskip]{3}{Water Potability} & task1 & Water Potability Chemistry Instability Analysis & \url{https://www.kaggle.com/datasets/adityakadiwal/water-potability} \\*
\cline{2-4}
  & task2 & Water Potability Drinking-Status Modeling & \url{https://www.kaggle.com/datasets/adityakadiwal/water-potability} \\*
\cline{2-4}
  & task3 & Water Potability Safety Screening and Model Comparison & \url{https://www.kaggle.com/datasets/adityakadiwal/water-potability} \\
\bottomrule
\end{tabular}
\end{table*}

\begin{table*}[p]
\centering
\scriptsize
\hypersetup{urlcolor=black}
\setlength{\tabcolsep}{2pt}
\renewcommand{\arraystretch}{1.03}
\caption{Kaggle Datasets and Notebooks in the Social Good Domain}
\label{tab:kaggle-sources-social-good}
\begin{tabular}{@{}K{0.24\textwidth}K{0.08\textwidth}K{0.31\textwidth}U{0.33\textwidth}@{}}
\toprule
\textbf{Competition/Dataset} & \textbf{Task ID} & \textbf{Notebook Title} & \textbf{URL} \\
\midrule
\SourceMultirow[-0.35\baselineskip]{2}{Data Science for Good: Center for Policing Equity} & task1 & Police Dogs and Grey hair will save you from jail & \url{https://www.kaggle.com/code/harriken/police-dogs-and-grey-hair-will-save-you-from-jail/report} \\*
\cline{2-4}
  & task2 & Very Detailed Analysis of CPE - DS for Good Winner & \url{https://www.kaggle.com/code/ambarish/very-detailed-analysis-of-cpe-ds-for-good-winner/comments?scriptVersionId=8384365} \\
\midrule
Data Science for Good: City of Los Angeles~\citep{data-science-for-good-city-of-los-angeles} & task1 & Phrasing: Improving Diversity Through Formatting & \url{https://www.kaggle.com/code/filthyilliterate/phrasing-improving-diversity-through-formatting/notebook} \\
\midrule
\SourceMultirow[-0.7\baselineskip]{3}{Data Science for Good: Kiva Crowdfunding} & task1 & Kiva Data Exploration & \url{https://www.kaggle.com/code/gpreda/kiva-data-exploration/report} \\*
\cline{2-4}
  & task2 & Simple Exploration Notebook - Kiva & \url{https://www.kaggle.com/code/sudalairajkumar/simple-exploration-notebook-kiva/notebook} \\*
\cline{2-4}
  & task3 & ExtenKiva Exploration - EDA & \url{https://www.kaggle.com/code/kabure/extenkiva-exploration-eda/notebook} \\
\midrule
\SourceMultirow[-0.35\baselineskip]{2}{CareerVillage.org~\citep{data-science-for-good-careervillage}} & task1 & Deepdive into careervillage & \url{https://www.kaggle.com/code/infocusp/deepdive-into-careervillage} \\*
\cline{2-4}
  & task2 & 'When I grow up I want to be .. ' & \url{https://www.kaggle.com/code/arjundas/when-i-grow-up-i-want-to-be\#Is-this-people-answers-specific-tags?} \\
\midrule
\SourceMultirow[-0.35\baselineskip]{2}{Data Science for Good: PASSNYC} & task1 & Target Schools \& Action Recommended for PASSNYC & \url{https://www.kaggle.com/code/laiyipeng/target-schools-action-recommended-for-passnyc?scriptVersionId=5753794} \\*
\cline{2-4}
  & task2 & Recommendations to PASSNYC based on Data Analysis & \url{https://www.kaggle.com/code/infocusp/recommendations-to-passnyc-based-on-data-analysis/notebook} \\
\bottomrule
\end{tabular}
\end{table*}

\begin{table*}[p]
\centering
\scriptsize
\hypersetup{urlcolor=black}
\setlength{\tabcolsep}{2pt}
\renewcommand{\arraystretch}{1.03}
\caption{Kaggle Datasets and Notebooks in the Sports Domain}
\label{tab:kaggle-sources-sports}
\begin{tabular}{@{}K{0.24\textwidth}K{0.08\textwidth}K{0.31\textwidth}U{0.33\textwidth}@{}}
\toprule
\textbf{Competition/Dataset} & \textbf{Task ID} & \textbf{Notebook Title} & \textbf{URL} \\
\midrule
Big Data Derby 2022~\citep{big-data-derby-2022} & task1 & Monitoring Racing Strategies for Injury Prevention & \url{https://www.kaggle.com/code/iamleonie/monitoring-racing-strategies-for-injury-prevention/} \\
\midrule
Google Cloud \& NCAA March Madness Analytics~\citep{march-madness-analytics-2020} & task1 & What Makes a Second-Half Team? & \url{https://www.kaggle.com/code/hmtessier/what-makes-a-second-half-team/notebook} \\
\midrule
NFL Big Data Bowl 2023~\citep{nfl-big-data-bowl-2023} & task1 & NFL Data Bowl 2023: Initial Pass Set Kick Speed & \url{https://www.kaggle.com/code/morganmartin23/nfl-data-bowl-2023-initial-pass-set-kick-speed} \\
\bottomrule
\end{tabular}
\end{table*}

\begin{table*}[p]
\centering
\scriptsize
\hypersetup{urlcolor=black}
\setlength{\tabcolsep}{2pt}
\renewcommand{\arraystretch}{1.03}
\caption{Kaggle Datasets and Notebooks in the Community Domain}
\label{tab:kaggle-sources-community}
\begin{tabular}{@{}K{0.24\textwidth}K{0.08\textwidth}K{0.31\textwidth}U{0.33\textwidth}@{}}
\toprule
\textbf{Competition/Dataset} & \textbf{Task ID} & \textbf{Notebook Title} & \textbf{URL} \\
\midrule
\SourceMultirow[-0.7\baselineskip]{3}{2018 Kaggle Machine Learning \& Data Science Survey} & task1 & AfricAI & \url{https://www.kaggle.com/code/mhajabri/africai} \\*
\cline{2-4}
  & task2 & The MOOC Wars: Kaggle's Perspective & \url{https://www.kaggle.com/code/ogakulov/the-mooc-wars-kaggle-s-perspective?scriptVersionId=8041710} \\*
\cline{2-4}
  & task3 & Measuring Accountability in DS and ML with Waffles & \url{https://www.kaggle.com/code/strangemane/measuring-accountability-in-ds-and-ml-with-waffles} \\
\midrule
\SourceMultirow[-0.7\baselineskip]{3}{2019 Kaggle Machine Learning \& Data Science Survey~\citep{kaggle-survey-2019}} & task1 & Exploring PhD Community with Network Analysis & \url{https://www.kaggle.com/code/artvolgin/exploring-phd-community-with-network-analysis} \\*
\cline{2-4}
  & task2 & Is there any job out there? Kaggle vs Glassdoor & \url{https://www.kaggle.com/code/andresionek/is-there-any-job-out-there-kaggle-vs-glassdoor/notebook} \\*
\cline{2-4}
  & task3 & Spending \$\$\$ for MS in Data Science - Worth it ? & \url{https://www.kaggle.com/code/shivamb/spending-for-ms-in-data-science-worth-it} \\
\midrule
\SourceMultirow[-0.7\baselineskip]{3}{2020 Kaggle Machine Learning \& Data Science Survey~\citep{kaggle-survey-2020}} & task1 & Treasure Hunt - what gives to be REALLY good? & \url{https://www.kaggle.com/code/andradaolteanu/treasure-hunt-what-gives-to-be-really-good} \\*
\cline{2-4}
  & task2 & Tools of the Trade: A Short History & \url{https://www.kaggle.com/code/haakakak/tools-of-the-trade-a-short-history/notebook} \\*
\cline{2-4}
  & task3 & How to make money in 2021 & \url{https://www.kaggle.com/code/viveknest/how-to-make-money-in-2021} \\
\midrule
\SourceMultirow[-0.7\baselineskip]{3}{2021 Kaggle Machine Learning \& Data Science Survey~\citep{kaggle-survey-2021}} & task1 & Data Science in 2021 : Adaptation or Adoption? & \url{https://www.kaggle.com/code/shivamb/data-science-in-2021-adaptation-or-adoption} \\*
\cline{2-4}
  & task2 & How are the Ladies and the Gents doing? & \url{https://www.kaggle.com/code/andradaolteanu/how-are-the-ladies-and-the-gents-doing} \\*
\cline{2-4}
  & task3 & Data Scientists \& Analysts: What's the difference? & \url{https://www.kaggle.com/code/spitfire2nd/data-scientists-analysts-what-s-the-difference/notebook} \\
\midrule
\SourceMultirow[-0.7\baselineskip]{3}{2022 Kaggle Machine Learning \& Data Science Survey~\citep{kaggle-survey-2022}} & task1 & 15 factors for data science in your country! & \url{https://www.kaggle.com/code/michau96/15-factors-for-data-science-in-your-country} \\*
\cline{2-4}
  & task2 & The State of Low / No-code in Data & \url{https://www.kaggle.com/code/spitfire2nd/the-state-of-low-no-code-in-data/} \\*
\cline{2-4}
  & task3 & Classifying Users and Learning From Experts & \url{https://www.kaggle.com/code/rosspmcdonald/classifying-users-and-learning-from-experts/notebook} \\
\midrule
GitHub Programming Languages Data & task1 & Data Visualization on Github Languages Data & \url{https://www.kaggle.com/code/varunnagpalspyz/data-visualization-on-github-languages-data} \\
\bottomrule
\end{tabular}
\end{table*}

\section{Prompt Templates}

\subsection{System Prompt}
\label{prompt:system_prompt}
\begin{tcolorbox}[
    colback=gray!10,
    colframe=black,
    title={System Prompt},
    breakable
]
\begin{lstlisting}[style=promptstyle]
You are an expert data scientist, statistical analyst and machine learning engineer who tackles analytical or machine learning challenges through systematic thinking and investigation.
For each task, you will receive a question along with file paths to the relevant data and background information in `{PATH}`. 
Your goal is to:
1. Understand the problem - interpret the question, data format, and expected output format.
2. Explore and preprocess the data - load the datasets, perform data cleaning, feature engineering, and exploratory analysis where helpful.
3. Decompose the question and perform planning - break down the question into smaller steps and perform each step systematically. Change your plan if needed.
4. Analyze the data - build appropriate statistical models, causal models, machine learning models, or other analyses to answer the research question.
5. Generate final answer - provide a clear, specific answer to the question based on your analysis and the requirements.
6. Explain reasoning - clearly communicate assumptions, methodology, and trade-offs at each step.

TASK: Tackle the given data science question by analyzing the provided data to generate a final answer.

Important rules:
- Do not use plotting libraries (assume you cannot view plots). Use text-based summaries and statistics instead.
- Your final answer should be specific and directly address the question.
- For numerical answers, provide the exact value requested (rounded as specified if mentioned).
- Only produce the final answer when you have enough evidence and validation to support your approach.
- Try different approaches or perform deeper reasoning when you are uncertain about the answer.
- Code execution is continuous - variables and data loaded in previous steps remain available for subsequent analysis. Do not need to reload the same dataset or variables.
- Your code can only do one step at a time even when multiple steps are planned. Perform the next step based on the previous step's results.
- When calculation is needed, you are encouraged to use python code instead of calculating by yourself.
- You must provide your final answer in the format: <answer>your final answer</answer>

You MUST use the following format for your response. Each step must follow this exact structure:
<reasoning>
Write clear reasoning about what you plan to do next and why. Be specific about your analytical approach.
</reasoning>
<python>
Write executable Python code here. Each code block should do ONE specific task.Code must be complete and runnable. Include all necessary imports.
</python>
<information>
The output/results from your Python code will appear here.\nThis section is read-only - you cannot write here.
</information>
Repeat these blocks for each analysis step. When you reach your conclusion, you should follow this structure:
<reasoning>
Write clear reasoning about how you came up with your final answer.
</reasoning>
<answer>
Write your final answer here according to the requirements of the question. Do not include any other text or unnecessary information.
</answer>


\end{lstlisting}
\end{tcolorbox}

\subsection{LLM-as-Judge Prompt}
\label{prompt:eval_prompt}
\begin{tcolorbox}[
    colback=gray!10,
    colframe=black,
    title={Eval Prompt},
    breakable
]

\begin{lstlisting}[style=promptstyle]
## Evaluation Task

You are a strict factual evaluator. Your job is to check whether an agent's solution correctly answers the question by verifying it against the relevant facts in
the ground truth.

---

### Inputs

**Question:**
{question}

**Ground Truth (JSON):**
{ground_truth}

**Agent's Solution:**
{solution}

---

### Evaluation Rules

1. **Question-Driven Coverage** - First, analyze the `question` to determine which specific information is requested. You ONLY need to evaluate the fields in the
`ground_truth` that directly answer the question. Ignore extra fields in the `ground_truth` that are not requested. Ignore extra information in the solution as
well, as long as all required information is present and correct. Missing required fields count as incorrect.

2. **Numeric values** - Numeric answers must match the ground truth exactly after ignoring insignificant trailing zeros.

- Compare numeric values exactly after normalizing trailing zeros after the decimal point.
- Trailing zeros after the decimal point are insignificant and should be ignored.
- A decimal point followed only by zeros is equivalent to an integer.
- Do NOT round values.
- Do NOT allow +/- 1 tolerance in the last digit.
- Do NOT compare using fewer decimal places unless the removed digits are only trailing zeros.
- Percent signs, currency symbols, commas, and surrounding text may be ignored for parsing, but the numeric value itself must still match exactly after trailing-zero normalization.

Examples:
- Ground Truth `22245.00` vs Solution `22245` -> Match
- Ground Truth `25.7600` vs Solution `25.76` -> Match
- Ground Truth `0.125` vs Solution `0.12` -> Wrong, numeric value differs

If the ground truth explicitly includes a `tolerance` or `tolerance_note` field for a required numeric value, apply that tolerance only to the numeric value. Trailing zeros may still be ignored unless the tolerance note explicitly requires fixed formatting.

3. **Numeric tolerance** - If the ground truth explicitly includes a `tolerance` or `tolerance_note` field for a required numeric value:
- Apply that tolerance **only** to the numeric value.
- Trailing zeros may still be ignored unless the tolerance note explicitly requires fixed formatting.

4. **Rankings / ordered lists** - Verify both the items and their order. **Exception for ties:** If multiple items have the exact same numerical value, any order
among those tied items is acceptable. Only evaluate rankings if the question actually asks for them.

5. **Label normalization / aliases** - Ignore differences in labels entirely. Do **not** consider variations in case, punctuation, spacing, apostrophes, typography, or shorthand forms when judging correctness. Label names are **not** used as a criterion for correctness; only the associated values or required information are evaluated.

6. **Formatting** - Ignore differences in wording, formatting, currency symbols, percent signs, or extra explanation. Judge factual correctness only.

7. **Scoring is binary** - Score **1** only if ALL required fields are correct. Score **0** if ANY required field is wrong or missing.

---

### Output Format

Reply in EXACTLY this format:

<reasoning>
Step 1: Identify which fields in the ground truth are actually requested by the question.
Step 2: Brief analysis of each required ground truth field vs. the solution. For numeric values, verify exact numeric equality after ignoring insignificant trailing zeros, with no rounding unless an explicit tolerance is provided. Apply label normalization for obvious aliases, and allow flexible ordering only for tied ranking values.
</reasoning>
<error>if Score is 0, list each incorrect or missing REQUIRED field and explain why it is wrong; if Score is 1, write "None"</error>
<score>0 or 1</score>
"""

\end{lstlisting}
\end{tcolorbox}

\subsection{Error Annotation Prompt}
\label{prompt:error_prompt}
\begin{tcolorbox}[
    colback=gray!10,
    colframe=black,
    title={Error Annotation Prompt},
    breakable
]
\begin{lstlisting}[style=promptstyle]
You are an expert in error analysis for multi-turn, long-horizon data analysis benchmarks. Your task is to analyze why the agent answered each task incorrectly, and to distinguish ordinary data-analysis errors from long-horizon errors.

Read and use these files together:
- `code.py`: code generated by the agent.
- `ground_code.py`: reference solution code.
- `results_eval.json`: contains each task's question, context, reference answer, agent answer, agent trajectory, and evaluation process. The field `success` indicates whether the agent produced an answer; `judge.score` indicates whether the answer is correct.
- `task.ipynb`: benchmark notebook with task order, context, constraints, and reference solution logic.
- If necessary, inspect actual data files, but do not reinvent the task meaning. Treat `task.ipynb` and `ground_code.py` as authoritative.

Goal:
For each incorrect task, determine:
1. What went wrong;
2. Why it went wrong;
3. Whether the error was newly introduced in the current task or propagated from previous tasks;
4. Whether the error is an ordinary data-analysis error or a long-horizon / multi-round error.

Analyze tasks where `judge.score == 0` as incorrect. Do not rely on `success` alone: `success=True` can still be semantically wrong, and `success=False` often means the agent failed to produce a valid final answer.

Error types must be selected only from the following categories.

## Error Types

### 1. Statistical / Domain Reasoning Error

Errors caused by misunderstanding statistical concepts, metric semantics, dataset structure, field meanings, entity definitions, population scope, scale, or domain-specific analytical assumptions.

Use this type when the agent's mistake comes from an incorrect understanding of what a metric, field, entity, threshold, population, or domain concept means, even if the code runs.

Examples:
- Misunderstanding percentile vs quantile, percentile rank vs raw quantile cutoff, average-rank percentile vs min-rank percentile, rank vs score, z-score, residual, shrinkage, weighted mean, or support;
- Misunderstanding whether a threshold is absolute, relative, inclusive, exclusive, raw-scale, percentile-scale, rank-based, or population-relative;
- Computing a percentile, rank, cutoff, cap, median, or quantile over the wrong conceptual population, such as current candidates instead of the full working table, full data instead of current pool, or all players instead of side-specific players;
- Misunderstanding the meaning of exposure, coverage, risk, lift, residual, market lift, publisher support, neighbor support, contributor support, trend support, actionability, fragility, robustness, or reliability;
- Misunderstanding domain-specific fields such as season, region, station, district, surface, race distance, track condition, publisher family, work family, school group, cohort, market, or time window;
- Misunderstanding the unit of analysis, such as start vs horse, book vs work family, edition vs title, school vs district, station vs region, restaurant vs order, user vs interaction, player vs team, race vs season;
- Misunderstanding which population a metric is defined over, such as clean comparison starts, current warnings, reviewed group, candidate pool, active sample, strong finishers, reliable candidates, or final stable candidates;
- Misunderstanding the interpretation of a derived label, such as severe-stable, broader-dependent, actionable, repeated-stable, late-specific, fragile, robust, supported, current candidate, or reliable candidate;
- Misunderstanding the meaning of a comparison or diagnostic, such as treating a sensitivity check as a new default rule, treating a diagnostic support set as direct evidence, or interpreting a robustness count as another warning rule.

This type emphasizes that the agent misunderstood the statistical, semantic, or domain meaning of the task objects.

Important:
If the error involves the wrong metric population, wrong percentile/rank scale, wrong threshold semantics, wrong unit of analysis, or wrong interpretation of a derived label, include Statistical / Domain Reasoning Error in `all_error_types`, even if the task also involves instruction following, state management, or cascading.

For automated labeling, err on the side of including Statistical / Domain Reasoning Error in `all_error_types` when the mismatch involves percentile population, rank convention, score scale, cutoff population, unit of analysis, or derived-label semantics.

Do not use Statistical / Domain Reasoning Error when:
- The task explicitly stated a requirement and the agent simply omitted it without attempting it -> Instruction Following Error;
- The agent understood the concept but chose the wrong multi-step analytical route -> Planning Error;
- The agent understood the target concept and route but made a concrete code bug -> Coding / Implementation Error;
- The agent used the wrong concrete prior artifact, candidate pool, branch, or score table -> State Management Error;
- The current task is locally reasonable but inherits an already-wrong upstream artifact -> Cascade Error.

### 2. Planning Error

Errors in task decomposition, analytical strategy, method selection, or reasoning path.

Examples:
- Failing to recognize that the task requires constructing a candidate pool before ranking;
- Treating a stratified analysis problem as a global average problem;
- Choosing the wrong statistical test, modeling method, or analytical route;
- Failing to decompose a multi-stage analysis task in the correct order;
- Misjudging which intermediate results or comparison targets are needed for the current task;
- Performing exploratory analysis instead of completing the requested analytical workflow.

This type emphasizes incorrect analytical route or method design.

If the model identified the correct route but implemented it incorrectly in code, label it as Coding / Implementation Error.
If the model failed to follow steps or conditions explicitly stated in the current task, label it as Instruction Following Error.
If the model used the wrong prior state or branch, label it as State Management Error.

### 3. Instruction Following Error

Failure to follow requirements explicitly stated in the current task, including output format, processing method, formula, filtering condition, sorting rule, comparison target, or analysis step.

The key criterion:
The requirement is directly stated in the current task text or current context, and the model should have followed it in this turn.

Examples:
- The current task explicitly requires certain output fields, but the agent omits them;
- The current task explicitly requires a Boolean conclusion, but the agent does not output it;
- The current task explicitly requires a specific Top-K size, but the count is wrong;
- The current task explicitly specifies a sorting rule, but the agent does not follow it;
- The current task explicitly specifies decimal precision, but the agent does not follow it;
- The current task explicitly gives a formula, but the agent uses another formula;
- The current task explicitly requires a filtering condition, comparison target, or analysis step, but the agent does not apply it;
- The agent returns an empty answer, off-topic answer, exploratory notes, or asks what to do next instead of answering the task.

Do not label an error as Instruction Following Error merely because the final result violates the task. Decide whether the model ignored the requirement, implemented it incorrectly, planned the analysis incorrectly, misunderstood the metric/domain semantics, or used the wrong prior state.

If the current task explicitly states the population for a percentile, rank, cap, cutoff, or score and the agent uses a different population, this is both Instruction Following Error and Statistical / Domain Reasoning Error.

If a rule was established in a previous task or global convention but is not restated in the current task, forgetting it is Context Memory Error.

If the current task explicitly requires using a previous state, candidate pool, reviewed group, rollback branch, or previous artifact, but the model uses the wrong one, prioritize State Management Error.

If the current response is empty or off-topic, primary_error_type should usually be Instruction Following Error unless the trajectory clearly shows a runtime/code failure.

### 4. Coding / Implementation Error

Errors caused by concrete code-level implementation mistakes after the agent has otherwise identified the correct task requirement, analytical route, data scope, and inherited state.

Use this type only when the intended logic is substantially correct, but the code implementation is wrong.

Examples:
- Syntax error, runtime error, missing import, undefined variable, or invalid column reference;
- Code references the wrong DataFrame or column despite using the correct intended artifact;
- Merge/join is attempted with the correct tables but uses the wrong key or join type;
- Groupby/aggregation is attempted at the correct conceptual level but implemented with wrong grouping columns;
- Filter is attempted for the correct condition but implemented with an incorrect operator, boundary, or boolean combination;
- Rank/sort is attempted on the correct metric and population but implemented with wrong ascending direction, rank method, or tie-breaker;
- Deduplication is attempted for the correct entity level but implemented with the wrong subset/order;
- Text normalization or Unicode handling is implemented incorrectly;
- A specified formula is recognized and attempted, but transcribed incorrectly in code;
- Model or nearest-neighbor pipeline is conceptually appropriate, but has a concrete implementation bug such as failing to exclude self-neighbors, using the wrong fitted transformer, or applying train/test split incorrectly.

This type emphasizes: correct intent, wrong code.

Do not use Coding / Implementation Error when:
- The model misunderstood the statistical or domain concept -> Statistical / Domain Reasoning Error;
- The model chose the wrong analysis strategy or comparison design -> Planning Error;
- The model ignored a requirement explicitly stated in the current task -> Instruction Following Error;
- The model forgot a rule established earlier -> Context Memory Error;
- The model used the wrong candidate pool, state, branch, score table, or previous artifact -> State Management Error;
- The current task is locally reasonable but inherits an already-wrong upstream artifact -> Cascade Error.

Do not label a task as Coding / Implementation Error merely because `code.py` differs from `ground_code.py`. First identify whether the difference is caused by wrong intent, wrong state, wrong metric semantics, or wrong implementation.

### 5. Context Memory Error

Failure to remember long-range rules, global conventions, or default specifications established earlier but not explicitly restated in the current task.

The key criterion:
The requirement is not newly given in the current task. It was established in an earlier task, global instruction, or earlier context, and the current turn requires long-horizon memory to continue following it.

Examples:
- Forgetting previously specified decimal precision;
- Forgetting a default sorting convention, such as "results are listed from strongest to weakest unless otherwise specified";
- Forgetting previously specified output field requirements;
- Forgetting a previously defined processing method, formula, filtering condition, or analysis step;
- Forgetting a previously defined naming convention, grouping rule, default filtering rule, or percentile convention;
- Forgetting the rule meaning of a previously defined reviewed group, candidate definition, or analysis convention.

This type emphasizes loss of rule memory.

Important boundary:
Context Memory Error concerns forgetting rules, conventions, definitions, or default specifications.
State Management Error concerns using the wrong concrete intermediate state, candidate pool, score table, branch, or artifact.
Statistical / Domain Reasoning Error concerns misunderstanding the meaning of a metric, population, threshold, or domain concept.

Do not use Context Memory Error as primary merely because an upstream score drifted. If the current task correctly uses an already-wrong upstream score/table/set, primary_error_type should be Cascade Error.

If the current task explicitly restates the requirement and the model still fails to follow it, prioritize Instruction Following Error.

### 6. State Management Error

Failure to correctly inherit, update, roll back, isolate, or reuse intermediate state from previous tasks.

Examples:
- Using the wrong current candidate pool;
- Using the wrong reviewed group, final set, lifted slice, or long-risk set;
- Using the wrong adjusted score, current quality, folded-market score, contributor-adjusted score, or other intermediate column;
- Continuing to use the current branch when the task requires the rollback state;
- Comparing the current result to raw data when the task requires comparing pre-rollback and post-rollback states;
- Letting a local perturbation leak into later default state;
- Continuing to use an old state when an updated definition should be inherited;
- Mixing separated-market treatment with merged-market treatment;
- Using the wrong previous model output or previous filtering result;
- Selecting the wrong saved artifact among several previous views, such as earliest formula, current score, middle-stage score, diagnostic score, or final score.

This type emphasizes using, inheriting, updating, rolling back, or isolating dynamic intermediate state incorrectly.

If the problematic object is a specific set, table, score, branch, candidate pool, model output, or intermediate result, prioritize State Management Error over Context Memory Error.

If the intended prior artifact exists and the current task selects the wrong one, this is State Management Error, not Cascade Error.

### 7. Cascade Error

The current task is locally reasonable under the inherited intermediate result, state, or semantic assumption, but that inherited artifact was already wrong due to an earlier task, causing the current answer to be wrong even without a new independent error.

Use Cascade Error only when:
- The current task has no obvious new independent error;
- The current task actually uses an incorrect upstream artifact, such as a wrong candidate pool, score, label, baseline, mapping, model output, or intermediate result;
- If the upstream artifact were corrected, the current answer would likely become correct or close to correct.

Do not mark Cascade Error merely because an earlier task was wrong.

Use Cascade Error as `primary_error_type` when the current task is locally reasonable and the main reason it is wrong is that it inherited an already-wrong artifact. In that case, do not use Context Memory Error or State Management Error as primary unless the current task independently forgot a rule or selected the wrong artifact.

Do not output `all_error_types = ["Cascade Error"]` unless the current task's procedure, metric semantics, population choice, state selection, and output format are all locally correct under the inherited artifact.

Examples:
- The current task correctly sums a score share, but the inherited score values from an earlier task are wrong.
- The current task correctly ranks the current warning set, but the current warning set was already wrong.
- The current task correctly decomposes the selected candidate, but the selected candidate came from an earlier wrong ranking.
- The current task correctly compares groups, but the group labels were already wrong upstream.

If the current task introduces a clear new error:
- `primary_error_type` should be the current-turn error type;
- `all_error_types` may include Cascade Error;
- `upstream_error_task_id` should identify the relevant upstream task.

If the current response is empty, off-topic, exploratory, or fails to answer the current task's explicit request, prioritize Instruction Following Error and do not mark Cascade Error unless the response clearly uses a specific incorrect inherited artifact.

## Annotation Rules

For each incorrect task:
- `primary_error_type`: choose exactly one type that most directly explains the current incorrect answer.
- `all_error_types`: include all relevant error types. Include Cascade Error only if the current task actually depends on an incorrect upstream artifact.
- `upstream_error_task_id`: use `null` if there is no cascade. If Cascade Error appears in `primary_error_type` or `all_error_types`, provide the upstream task id; use an array if multiple tasks contributed.
- If the current task has a new independent error, do not use Cascade Error as the primary type.
- If the current task is locally correct or mostly reasonable but uses an already-wrong upstream artifact, use Cascade Error as primary.
- If the correct prior artifact exists but the current task selects, updates, rolls back, or reuses the wrong artifact, use State Management Error as primary.
- If the current task uses the intended prior artifact correctly but that artifact was already wrong, use Cascade Error as primary.
- If the agent used the wrong data scope, candidate pool, prior state, branch, or intermediate table, prefer State Management Error over Coding / Implementation Error.
- If the agent used a different formula because it ignored the formula explicitly stated in the current task, prefer Instruction Following Error.
- If the agent used a different formula because it misunderstood the metric meaning, scale, threshold, or population, prefer Statistical / Domain Reasoning Error.
- If the agent selected the wrong analytical route before coding, prefer Planning Error.
- If the error involves percentile/rank population, cutoff population, metric scale, threshold semantics, entity granularity, or unit of analysis, include Statistical / Domain Reasoning Error in `all_error_types`.
- Before assigning Cascade Error, explicitly check whether the current task itself introduced any independent error.
- If the current task explicitly states a formula, population, percentile ladder, candidate pool, ranking scope, comparison target, or output requirement, and the agent violates it, then `all_error_types` must include Instruction Following Error, even if the task also inherits an upstream error.
- If the current task involves percentile, rank, cutoff, cap, threshold, population semantics and the agent uses the wrong population, scale, rank convention, or threshold interpretation, then `all_error_types` must include Statistical / Domain Reasoning Error.
- Cascade Error alone is allowed only when the current task has no independent mismatch with the current task instruction, metric semantics, population choice, state selection, or output format.
- If the current task both inherits a wrong upstream artifact and violates a current-task requirement, `all_error_types` must include both Cascade Error and the current-task error type.
- If the current task uses a wrong score, label, candidate pool, or table that was produced by a previous incorrect task, `all_error_types` must include Cascade Error and `upstream_error_task_id` must name that previous task.
- If the current task selects the wrong prior artifact among multiple available prior artifacts, `all_error_types` must include State Management Error. If the selected artifact was also already wrong, `all_error_types` may also include Cascade Error.
- Do not let an empty or off-topic answer in an upstream task automatically become the root cause for later tasks. For downstream tasks, identify the actual wrong artifact being inherited.
- Do not truncate any field with ellipses such as "...". Write complete concise sentences.

## Analysis Steps

1. Read `results_eval.json` first and identify all tasks with `judge.score == 0`.
2. For each incorrect task, inspect the current question, context, reference answer, agent answer, trajectory, and evaluation reasoning.
3. Compare `ground_code.py` and `code.py` to determine whether the discrepancy comes from output format, filtering scope, sorting, formula, join, aggregation, modeling pipeline, state inheritance, context memory, statistical/domain semantics, or upstream cascading.
4. If needed, review previous tasks in `task.ipynb` to determine whether a missed condition was stated in the current task or only established earlier.
5. If the task depends on a previous result, check whether the previous artifact was already wrong and whether the current task actually used it.
6. Explain the concrete mechanism causing the mismatch. Do not infer only from final numerical differences.
7. When multiple error types overlap, choose the most direct current cause as `primary_error_type` and include other relevant types in `all_error_types`.

## Required Output

Return a valid JSON object only. Do not output Markdown or explanatory prose.

Use this format:

{
  "summary": {
    "total_incorrect_tasks": 0,
    "primary_error_type_counts": {
      "Statistical / Domain Reasoning Error": 0,
      "Planning Error": 0,
      "Instruction Following Error": 0,
      "Coding / Implementation Error": 0,
      "Context Memory Error": 0,
      "State Management Error": 0,
      "Cascade Error": 0
    },
    "all_error_type_counts": {
      "Statistical / Domain Reasoning Error": 0,
      "Planning Error": 0,
      "Instruction Following Error": 0,
      "Coding / Implementation Error": 0,
      "Context Memory Error": 0,
      "State Management Error": 0,
      "Cascade Error": 0
    },
    "cascade_error_count": 0,
    "main_failure_patterns": [
      "Main failure pattern 1",
      "Main failure pattern 2"
    ]
  },
  "task_error_analysis": [
    {
      "task_id": 1,
      "primary_error_type": "one of the allowed error types",
      "all_error_types": ["one or more allowed error types"],
      "upstream_error_task_id": null,
      "error_summary": "Complete concise explanation of why the task is wrong.",
      "evidence": "One concise sentence comparing the task/reference behavior with the agent behavior.",
      "confidence": "high / medium / low"
    }
  ]
}

Save the final JSON object to a file named `error_ana.json` in the current working directory.
\end{lstlisting}
\end{tcolorbox}

\end{document}